\documentclass[]{fairmeta}

\usepackage{array}
\usepackage{xcolor}
\usepackage{graphicx}
\usepackage{float}
\usepackage{paralist}
\usepackage{bbding}

\usepackage{graphicx}
\usepackage{caption}
\usepackage{stfloats} 
\usepackage{amssymb}

\usepackage{hyperref}
\usepackage{url}
\usepackage{pifont}
\usepackage[inline]{enumitem}
\newcommand{\xmark}{\ding{55}}%
\usepackage{balance} 

\newcommand{\BibTeX}{\rm B\kern-.05em{\sc i\kern-.025em b}\kern-.08em\TeX}
\newcommand{\mlebench}{MLE-bench}

\title{What Does It Take to Be a Good AI Research Agent? \\ Studying the Role of Ideation Diversity \\}

\author[1]{Alexis Audran-Reiss}
\author[1]{Jordi Armengol-Estapé}
\author[1,2]{Karen Hambardzumyan}
\author[1]{Amar Budhiraja}
\author[1]{Martin Josifoski}
\author[1,2]{Edan Toledo}
\author[1]{Rishi Hazra}
\author[1]{Despoina Magka}
\author[1]{Michael Shvartsman}
\author[1]{Parth Pathak}
\author[1]{Justine T Kao}
\author[1]{Lucia Cipolina-Kun}
\author[1]{Bhavul Gauri}
\author[1]{Jean-Christophe Gagnon-Audet}
\author[1]{Emanuel Tewolde}
\author[3,4]{Jenny Zhang}
\author[1]{Taco Cohen}
\author[1]{Yossi Adi}
\author[1]{Tatiana Shavrina}
\author[1]{Yoram Bachrach}

\affiliation[1]{FAIR at Meta}
\affiliation[2]{University College London}
\affiliation[3]{Meta SuperIntelligence Labs}
\affiliation[4]{University of British Columbia}

\abstract{AI research agents offer the promise to accelerate scientific progress by automating the design, implementation, and training of machine learning models. However, the field is still in its infancy, and the key factors driving the success or failure of agent trajectories are not fully understood. We examine the role that ideation diversity plays in agent performance. First, we analyse agent trajectories on MLE-bench, a well-known benchmark to evaluate AI research agents, across different models and agent scaffolds. Our analysis reveals that different models and agent scaffolds yield varying degrees of ideation diversity, and that higher-performing agents tend to have increased ideation diversity. Further, we run a controlled experiment where we modify the degree of ideation diversity, demonstrating that higher ideation diversity results in stronger performance. Finally, we strengthen our results by examining additional evaluation metrics beyond the standard medal-based scoring of MLE-bench, showing that our findings still hold across other agent performance metrics.}

\date{November 20, 2025}
\correspondence{Alexis Audran-Reiss at \email{a.audranreiss@meta.com}, Jordi Armengol-Estapé at \email{jordiae@meta.com}}

\begin{document}

\maketitle

\section{Introduction}
\label{sec:intro}

The rapid advancement of Large Language Model-based~\citep{NEURIPS2020_1457c0d6} agents equipped with tools~\citep{schick2023toolformer} has sparked interest in the quest to develop research agents, in areas as challenging  as chemistry~\citep{Boiko2023} or biology~\citep{Swanson2025}.

\begin{figure}[H]
  \centering
    \centering
    \includegraphics[width=0.47\textwidth]{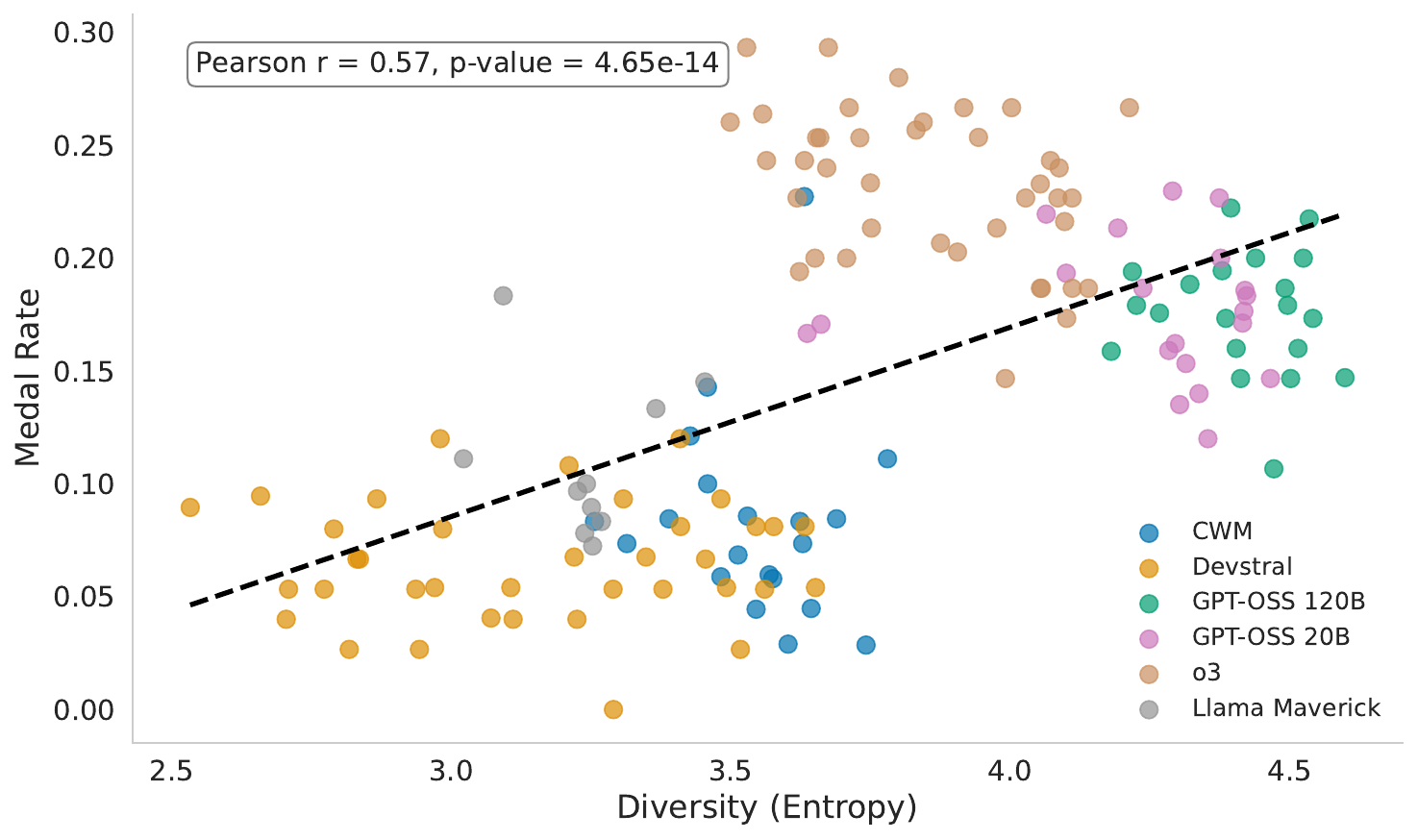}
    \caption{Ideation diversity correlates with performance in \mlebench: Our analysis shows that ideation diversity correlates with the agent's trajectory success in \mlebench. To confirm this relation, we later intervene on ideation diversity in a controlled experiment, in Section~\ref{sec:results_controlled_exp}.}
    \label{fig:correlationdiversityperformance_perseed}
\end{figure}

These research agents constitute an emerging paradigm in computational scientific discovery characterized by end-to-end autonomous systems capable of conducting independent research. 

In particular, recent work on autonomous AI research agents  \citep{shen2023hugginggptsolvingaitasks,huang2024mlagentbenchevaluatinglanguageagents, toledo2025airesearchagentsmachine, zhao2025automatedllmspeedrunningbenchmark} improves upon automated machine learning engineering tools \citep{autosklearn} by mirroring the cognitive processes of human researchers through a structured research pipeline: idea generation and hypothesis setup, experimental design and implementation, empirical validation, and iterative refinement. 
Recent advances have achieved notable milestones such as  creation of the first fully autonomous AI-generated research paper accepted through peer review \citep{yamada2025aiscientistv2workshoplevelautomated}.

Despite the potential of these recent breakthroughs in automating AI science, the field is still in its infancy and little is understood about the factors driving their successes and failures. Error analysis is substantially more complicated than in classic machine learning setups, due to the presence of long multi-step trajectories often guided by heuristic-based search algorithms~\citep{toledo2025airesearchagentsmachine} and leveraging tool use, which requires complex evaluation frameworks. Moreover, obtaining large-enough samples to perform meaningful analysis and ablate design choices can be computationally prohibitive.

This paper starts from the postulate that \textit{ideation diversity} is a key bottleneck in AI research agents' performance. To study this hypothesis, we face two key challenges: analyzing complex agentic trajectories at scale, and measuring and controlling ideation diversity. 

We perform a first-of-its-kind, large-scale study of AI research agents' trajectories in \mlebench~\citep{chan2025mlebench}, a well-known benchmark of Kaggle machine learning tasks. We study 6 different LLM backbones equipped with 2 different agentic frameworks (or \textit{scaffolds}) on the 75 machine learning tasks available in \mlebench \space across 10 to 20 random seeds, yielding a total of 11,000 trajectories. This corresponds to roughly 1,200,000 individual nodes in the agent scaffold search, for a total of 264,000 GPU hours.

To measure ideation diversity, we propose calculating Shannon entropy \citep{shannon1948mathematical} on the distribution of model architectures that the agent plans to implement in the ideation phase.  
Figure~\ref{fig:correlationdiversityperformance_perseed} shows the correlation between ideation diversity and the performance in \mlebench \space using our generated trajectory bank, hinting at a relation between the two.  

To confirm the diversity hypothesis, we perform a controlled experiment where we remove mechanisms yielding highly diverse solutions by amending the prompt.\footnote{Our main results in Section \ref{sec:results_controlled_exp} are based on decreasing ideation diversity through the prompt shown to the agent. In the Appendix \ref{sec:appendix}, we provide additional results where we control diversity via the sampling temperature parameter.} We run the controlled experiment on a subset of \mlebench, studying two diversity setups (control and ablated) using two agentic frameworks on 22 machine learning tasks across 10 seeds. We show that when ablating ideation diversity, the agents' performance decreases.  Finally, we strengthen this finding  by examining additional evaluation metrics on \mlebench \space aside from the standard score based on the Kaggle medal system.

\subsection{Contributions}

In summary, the contributions of this paper are as follows:
\begin{itemize}
\item We propose methods to \textbf{quantify and control the agent's ideation diversity}.
\item We perform a \textbf{first-of-its-kind, large-scale analysis on agentic trajectories}. We study a total of 11,000 AI research agents' trajectories across multiple agentic frameworks, LLM backbones, and machine learning tasks.
\item We show that the choice of \textbf{agentic scaffold significantly influences ideation diversity}. Our results further reveal a significant correlation between ideation diversity and agent performance on \mlebench \space tasks.
\item Through a \textbf{controlled experimental design}, we establish a causal relationship, showing that \textbf{increasing ideation diversity leads to improved performance} on \mlebench \space tasks.
\item We confirm that these findings are \textbf{robust when evaluated with alternative performance metrics}.
\end{itemize}

\section{Research Agents and \mlebench}

\paragraph{Agents} Broadly, \citet{soton252102} define an agent as a 
\textit{computer system that is situated in some environment and that is capable of autonomous action in this environment in order to meet its design objectives.} 
In the particular context of recent work on research agents based on generative AI and LLMs \citep{schick2023toolformer, Boiko2023, shen2023hugginggptsolvingaitasks,huang2024mlagentbenchevaluatinglanguageagents, Swanson2025, toledo2025airesearchagentsmachine, zhao2025automatedllmspeedrunningbenchmark}, we specifically refer to agent systems implemented using two main components: \begin{enumerate*} \item a model backbone (typically, an LLM), which processes observations from the environment as prompts and emits text-based actions, and \item  an outer loop making use of the model backbone to interface with the environment\end{enumerate*}.  This outer loop orchestrating the LLM actions is usually referred to as \textit{agentic frameworks} or \textit{agentic scaffolds} in the literature \citep{wu22}. The environments for AI research agents allow the agent to create and run code among other \textit{tools} \citep{schick2023toolformer}.

\label{sec:background}
\begin{figure*}[!htb]
    \centering
    \includegraphics[width=0.99\linewidth]{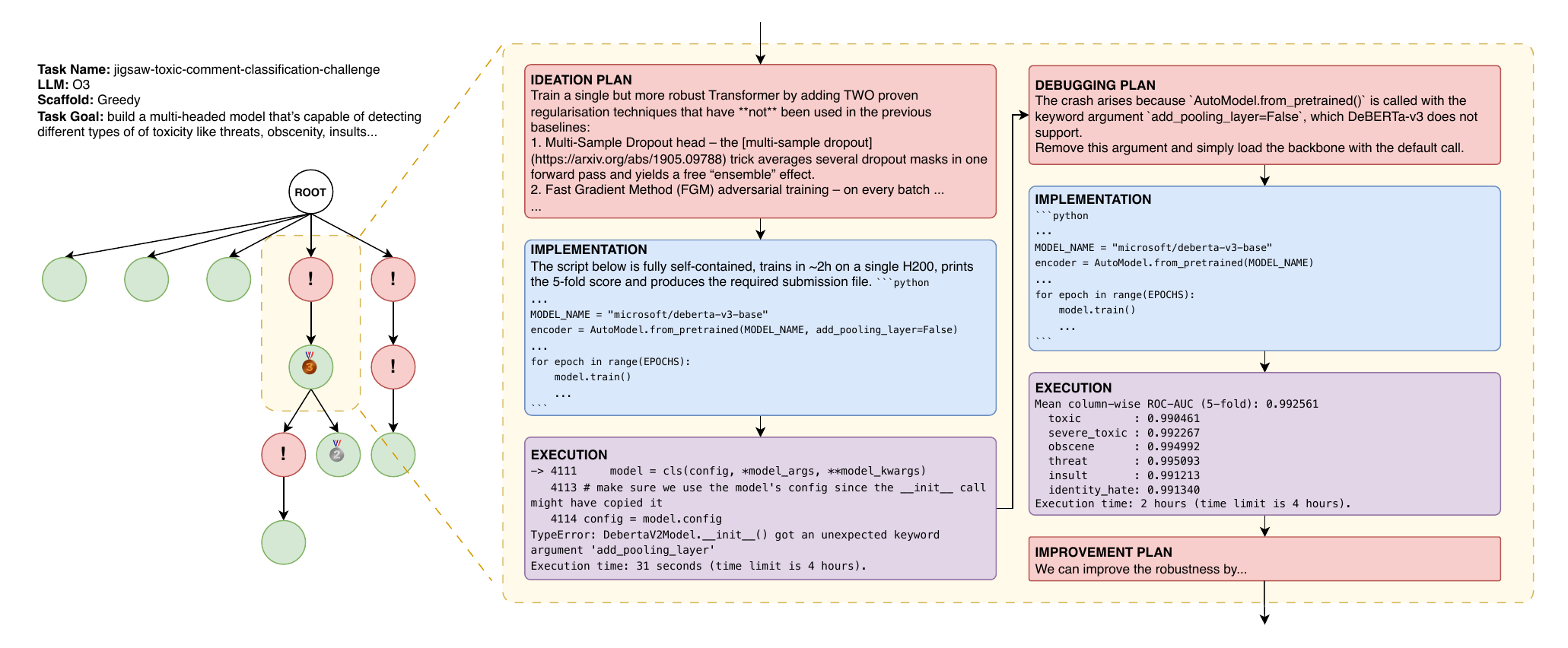}
    \caption{An example flow of an AI research agent attempting an \mlebench \space task. The goal of the task is to build a multi-headed model to classify different types of toxicity threats. The agent first tries the idea to finetune a model end to end, but the code fails and the agent fixes the bug. After analysis of this approach, the agent continues improving the solution, producing more nodes.
    }
    \label{fig:research_workflow}
\end{figure*}

\paragraph{MLE-Bench}~\citep{chan2025mlebench} (Machine Learning Engineering Benchmark) is a well-known evaluation framework designed to assess autonomous AI agents' capabilities in solving real-world machine learning problems. The benchmark is constructed from 75 tasks sourced from Kaggle competition datasets, providing a diverse collection of machine learning challenges that span computer vision, natural language processing, time series forecasting, tabular data analysis, and multimodal learning domains.

Each \mlebench \space task comprises standardized components including problem documentation, training datasets, held-out test sets, sample submission formats, and automated evaluation protocols. The evaluation methodology follows Kaggle's competitive framework, utilizing problem-specific metrics such as accuracy, F1-score, or RMSE, with performance assessed through leaderboard ranking systems.

AI research agents interact with \mlebench \space through a standardized API that mirrors real-world ML development workflows. Agents must autonomously perform the complete machine learning pipeline: data exploration, feature engineering, model selection, hyperparameter optimization, and submission generation within computational constraints. The benchmark employs stratified sampling and cross-validation methodologies to ensure robust performance assessment, with final rankings determined through holdout test set evaluation to prevent overfitting to validation metrics.

Given the same dataset, there are multiple ways of defining how to interface with the benchmark, mediated by the agentic scaffold of choice. Recent work \citep{toledo2025airesearchagentsmachine} proposed  a tree-based agentic scaffold to tackle \mlebench. In Section \ref{sec:agentic_scaffolds}, we describe the scaffolds studied in this work with more detail.

Figure \ref{fig:research_workflow} shows an example of a tree-search-based AI research agent  attempting the \textit{Jigsaw Toxic Comment Classification Challenge} \mlebench \space task. The AI research agent would execute the following computational workflow: idea generation (e.g., leveraging pre-trained convolutional neural network~\citep{lecun98} features with linear classifiers), hypothesis setup (e.g., establishing baselines using CIFAR-100 \citep{krizhevsky2009learning} embeddings with logistic regression), and implementation and experimentation (e.g., tuning learning rate for logistics regression). The next step would be to analyze results (e.g. look at confusion matrix) and finally make a submission to the grader. Based on the leaderboard rank and experiment analysis, the agent would then propose the next set of ideas and hypotheses (e.g. changing classifier from logistic regression to random forest or tuning a CNN end-to-end). 

These idea generation, implementation, experimentation and submission steps  are iterated upon to improve the agent's leaderboard rank. This paper focuses on the idea generation step.

\section{Methods}
\label{sec:methods}

In this section, we briefly introduce the methodology (including data, metrics, agentic orchestrations or scaffolds, and LLM backbones) used in both our data analysis, in Section~\ref{sec:results_diversity}, and the controlled experiment, in Section~\ref{sec:results_controlled_exp}.

\subsection{General Setup}

\subsubsection{Dataset} For the trajectory analysis, we use agent trajectories on \mlebench. For the controlled experiment, we focus on \mlebench \space lite, a curated subset of 22 tasks selected from the full benchmark.

\subsubsection{Metrics}

In line with the benchmark guidelines, for both our data analysis and controlled experiment, we assess each agent’s performance using the Medal Success Rate (henceforth referred to as medal rate).
Specifically, for each task, agents earn a bronze, silverho, or gold medal according to task-specific percentile
thresholds. We report the percentage of attempts in which an agent secures a medal. Later, in Section~\ref{sec:results_metrics}, we incorporate additional metrics. 

\subsubsection{Agentic Scaffolds}
\label{sec:agentic_scaffolds}

Following recent work~\citep{toledo2025airesearchagentsmachine}, we formalize AI research agents as search algorithms composed of a search policy, used to navigate the space of candidate solutions to a task, and a set of operators, which modify existing solutions to generate new candidate solutions.

Enabled by this formalization, we study a range of agentic structures, specifically focusing on:
(1) AIDE~\citep{aide2025}, an LLM-driven agent that approaches problem-solving as a tree-search over the domain of Python solutions, utilizing a Greedy policy. (2) $\text{AIRA}_{\text{GREEDY}} $~\citep{toledo2025airesearchagentsmachine}, another greedy tree-based search policy, with a different design for operators, memory scope, and prompts, and (3) $ \text{AIRA}_{\text{MCTS}} $~\citep{toledo2025airesearchagentsmachine},  utilizing Monte Carlo Tree Search (MCTS~\citep{mcts1, mcts2, mcts3}) for its search policy, in contrast to its greedy counterparts. 

In all three agentic scaffolds, the process results in trees where each node represents a Python code solution, created by one of the following operators: \begin{enumerate*}
    
\item \textbf{Draft}, which generates the initial population of solutions; \item  \textbf{Debug}, which identifies and corrects errors within a given node; and \item  \textbf{Improve}, which enhances the solution of a given node to increase its performance according to evaluation criteria.\end{enumerate*}

Additionally, the memory configuration dictates how each operator is selectively provided with previously produced artifacts, with well-scoped memory preventing issues such as context overload, mode collapse, and debug loops.

\subsubsection{LLM Backbones} 

For the data analysis, we use the following LLM backbones for the agents studied: o3~\citep{jaech2024openai}, gpt-oss~\citep{openai2025gptoss120bgptoss20bmodel} (20B and 120B), Llama Maverick~\citep{meta2025llama4}, Devstral \citep{rastogi2025devstralfinetuninglanguagemodels} and CWM~\citep{cwm2025}. Those represent different model sizes and architectures.

We conduct the controlled experiment discussed in Section~\ref{sect:control_div} with the full-sized DeepSeek R1 model~\citep{deepseekai2025deepseekr1incentivizingreasoningcapability}. All of the LLMs  above use a 128K-token context window to ensure input coverage without truncation.

\subsection{Measuring Ideation Diversity} \label{sect:meas_div}

Diversity can manifest in many aspects of machine learning engineering, such as data preprocessing, feature engineering, model development, and validation.
In this analysis, our focus is limited to examining the diversity of machine learning models trained by agents.

All AI research agents in our study begin their exploration by generating at maximum five initial ideas to solve the task at hand (exactly five for greedy searches, and up to five for MCTS), using the \textsc{Draft} operator. To measure ideation diversity, we compare agents by extracting two pieces of information from these five initial ideas. First, we extract
    the high-level ML approach or architecture used by our agent (for example CNN \citep{lecun98}, Transformer \citep{vaswani2017attention}, Decision Trees);
    and second, we also extract the specific model employed by the agent, with variants grouped together (e.g., EfficientNet-B4 is grouped as EfficientNet \citep{tan2019efficientnet}).

We study whether the design of the agent has a significant impact on the diversity of ideas generated by comparing the distribution of models used by AI research agents.

To quantify diversity, we leverage the model architectures that the agent intends to train. From the distribution of model architectures, we compute the Shannon entropy (in base 2), quantifying the average uncertainty (and therefore diversity) of the model architecture used by the AI research agent.

\subsection{Controlling Ideation Diversity}  \label{sect:control_div}
As part of our experimentation, we control the level of diversity using the system prompt in the LLM behind the agent. We compare two levels of diversity, the baseline agents and the agents with ablated diversity.

\subsubsection{Baseline agents}

We run baseline (or control) agents (with 2 scaffolds, $ \text{AIRA}_{\text{Greedy}} $, and $ \text{AIRA}_{\text{MCTS}} $), which, by default, are equipped with three mechanisms to enhance diversity.
\begin{enumerate*}
\item \textit{Sibling memory}, which provides to a new draft node the memory of its siblings, by including in the context descriptions of the solutions devised by the sibling nodes.
\item  \textit{Prompt-adaptive complexity}, which is a dynamic complexity cue within the system prompt aiming to guide the complexity of artifacts generated by the agents. For the first initial idea, we ask the agent to come up with an idea of minimal complexity. For the next two initial ideas, the system prompt asks for moderate complexity, and advanced complexity for the last two initial ideas.
\item \textit{Mention of diversity in the system prompt}, asking the base model to come up with different aspects of the solution every time.
\end{enumerate*}

\subsubsection{Agents with ablated diversity}

We remove prompt-adaptive complexity and the mention of diversity in the system prompt and we reuse sibling memory to request from the agent in the system prompt to come up with similar ideas.

By changing the parts of the prompt mentioning diversity, we intend to only impact the diversity of ideas generated by the agent, and not other solution aspects, such as implementation quality.
\section{The Ideation Diversity Bottleneck}
This section presents the core results of this paper. First, in Section~\ref{sec:results_diversity}, we analyze a large sample of agentic trajectories in \mlebench.  In Section~\ref{sec:results_controlled_exp}, we show the controlled experiment to validate the diversity hypothesis. Finally, in Section~\ref{sec:results_metrics}, we incorporate additional metrics in our analysis.

\label{sec:results}
\subsection{Deep-dive on Agentic Trajectories}

\label{sec:results_diversity}

\begin{figure*}[htbp]
    \centering
    \begin{minipage}[t]{0.48\textwidth}
        \centering
        \includegraphics[width=\linewidth]{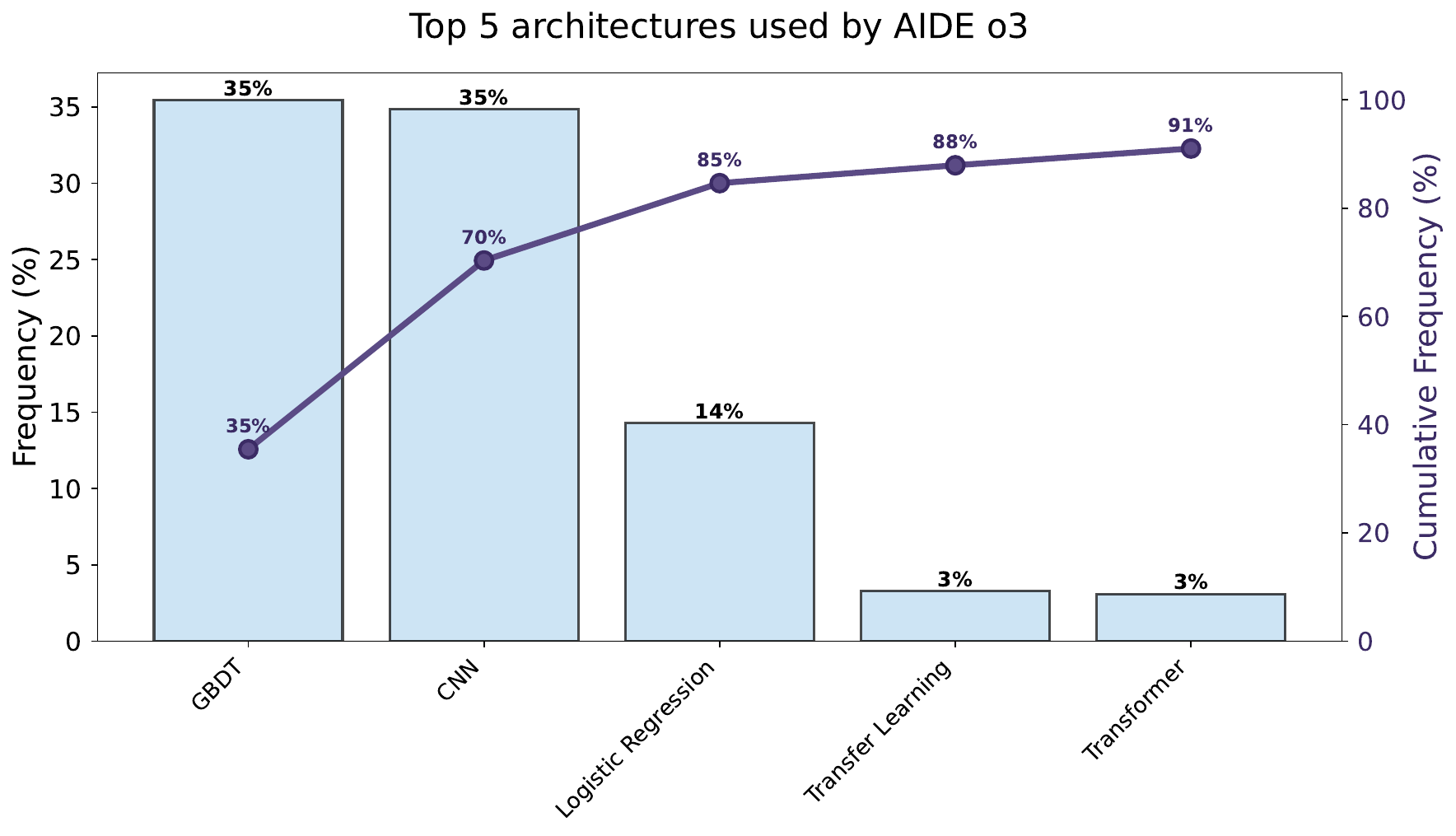}
        \caption*{(a) Diversity of ML approaches/architectures - AIDE}
    \end{minipage}%
    \hfill
    \begin{minipage}[t]{0.48\textwidth}
        \centering
        \includegraphics[width=\linewidth]{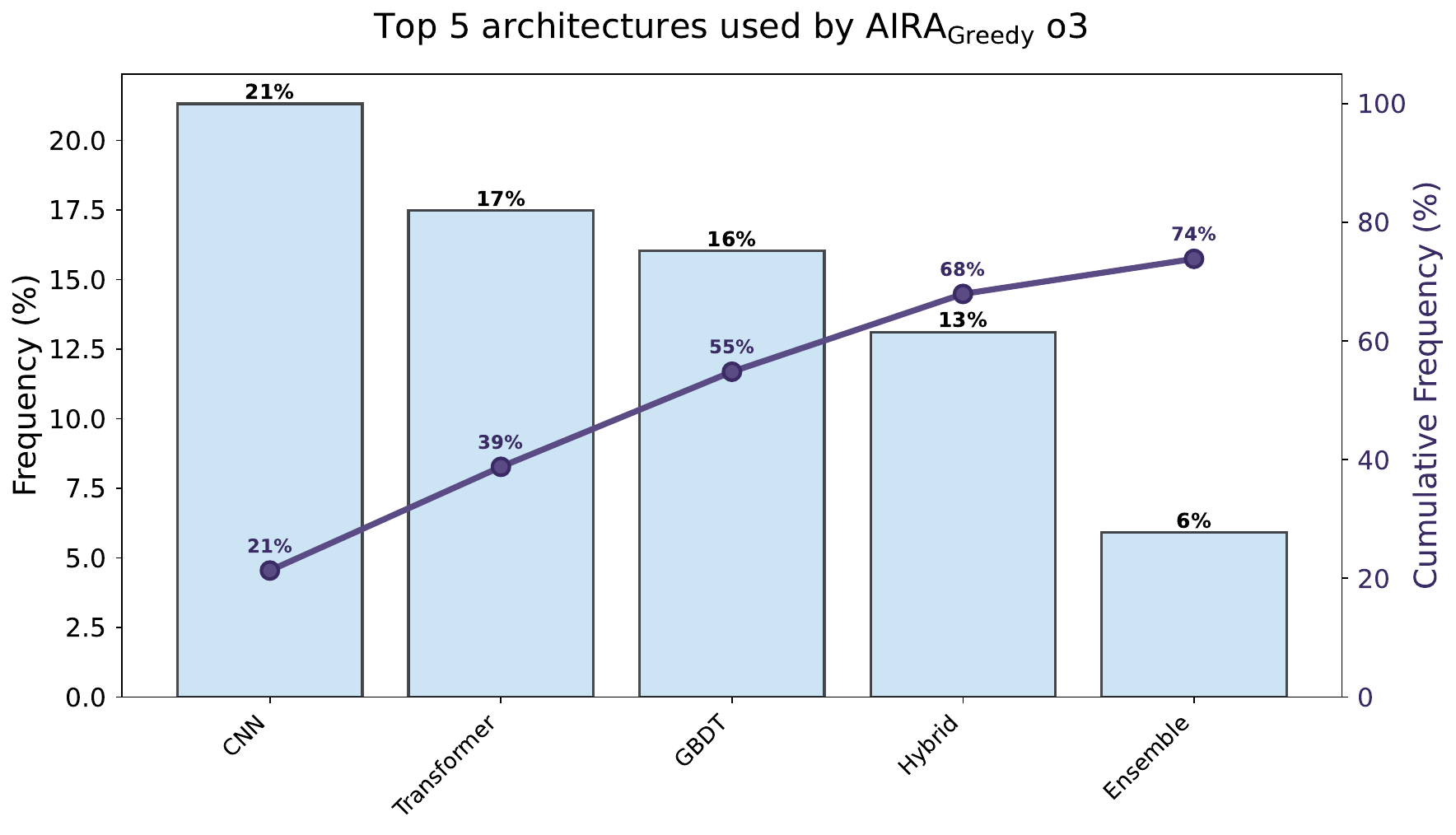}
        \caption*{(b) Diversity of ML approaches/architectures - $ \text{AIRA}_{\text{Greedy}} $}
    \end{minipage}

    \vspace{0.5cm}

    \begin{minipage}[t]{0.48\textwidth}
        \centering
        \includegraphics[width=\linewidth]{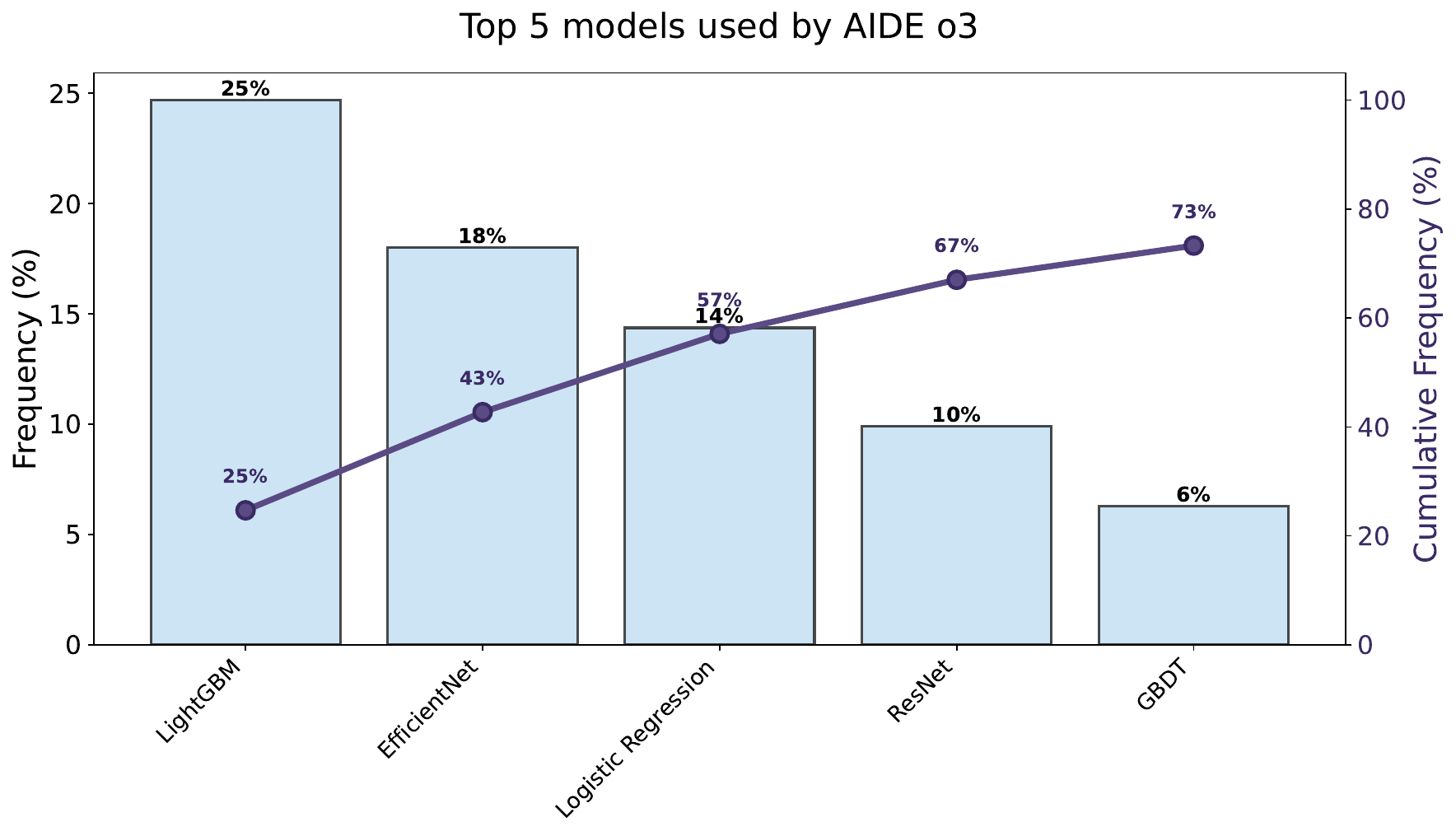}
        \caption*{(c) Diversity of models - AIDE}
    \end{minipage}%
    \hfill
    \begin{minipage}[t]{0.48\textwidth}
        \centering
        \includegraphics[width=\linewidth]{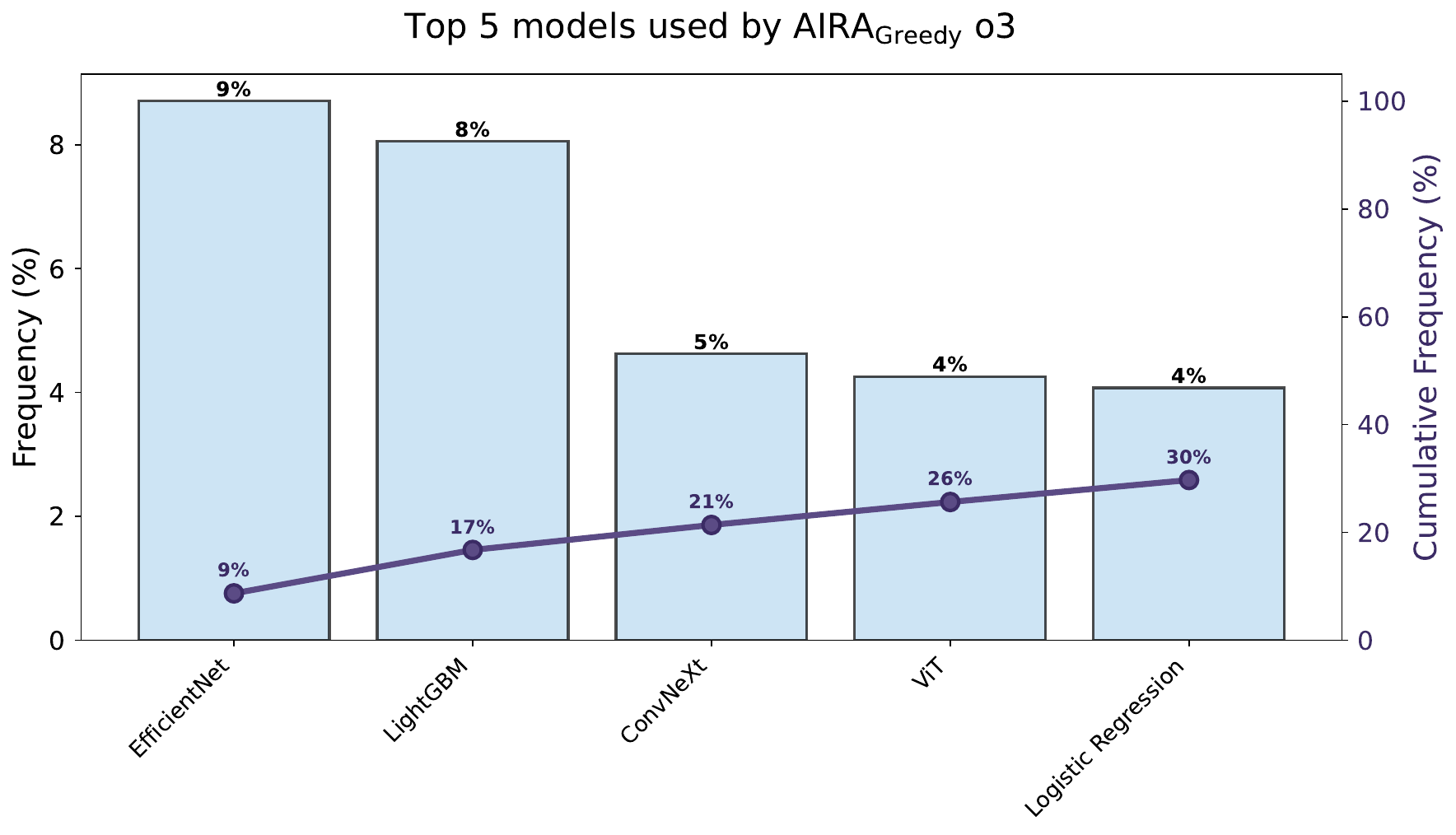}
        \caption*{(d) Diversity of models - $ \text{AIRA}_{\text{Greedy}} $}
    \end{minipage}

    \caption{Overview of diversity in models and architectures used on the 22 \mlebench \space lite tasks, illustrating the differences between the AIDE and $\text{AIRA}_{\text{Greedy}}$ scaffolds. (a, b) Distribution of architectures for AIDE and $\text{AIRA}_{\text{Greedy}}$.  
    (c, d) Distribution of model families.  
    }
    \label{fig:diversity_grid}
\end{figure*}

\paragraph{The agent scaffold choice impacts ideation diversity} Figure~\ref{fig:diversity_grid} illustrates a comparison between AIDE and $\text{AIRA}_\text{Greedy}$ agents, both using o3 as the backbone. Observing the model architectures and general machine learning approaches used by the agent, we can see from Figure~\ref{fig:diversity_grid}(a) that AIDE agents prefer Gradient Boosting Decision Trees (GBDT) and Convolutional Neural Networks (CNN) in 70\% of the initial draft nodes. 
In contrast, $\text{AIRA}_\text{Greedy}$ agents generate a greater diversity of ideas. The most common architectures among these agents are CNN, Transformers, GBDT, and Hybrid models that combine multiple approaches. Collectively, these four architectures represent 68\% of the ideas produced, as shown in Figure~\ref{fig:diversity_grid}(b).

Looking at the models trained by the agent in Figure~\ref{fig:diversity_grid}(c) and (d), LightGBM and EfficientNet represent 43\% of models AIDE agents intend to train in its initial draft nodes, while in the case of 
$\text{AIRA}_\text{Greedy}$ as many as 9 models represent this percentage. This difference in diversity highlights the importance of the design of agents (system prompt, search mechanism, operators), in influencing the variety of models the agent intends to train or use. Next, we study how this ideation diversity correlates to performance on \mlebench.

 \paragraph{Diversity correlates with performance on  \mlebench} 
Figure~\ref{fig:correlationdiversityperformance_perseed} shows the correlation of the \mlebench \space score (measured as medal rate) with ideation diversity (measured as architecture choice distribution entropy), for each run of the AI research agents included in our study on the 75 tasks of \mlebench. Here, a point in the plot refers to one agent's performance on the full set of tasks.

\begin{figure}[H]
    \centering
    \includegraphics[width=8cm]{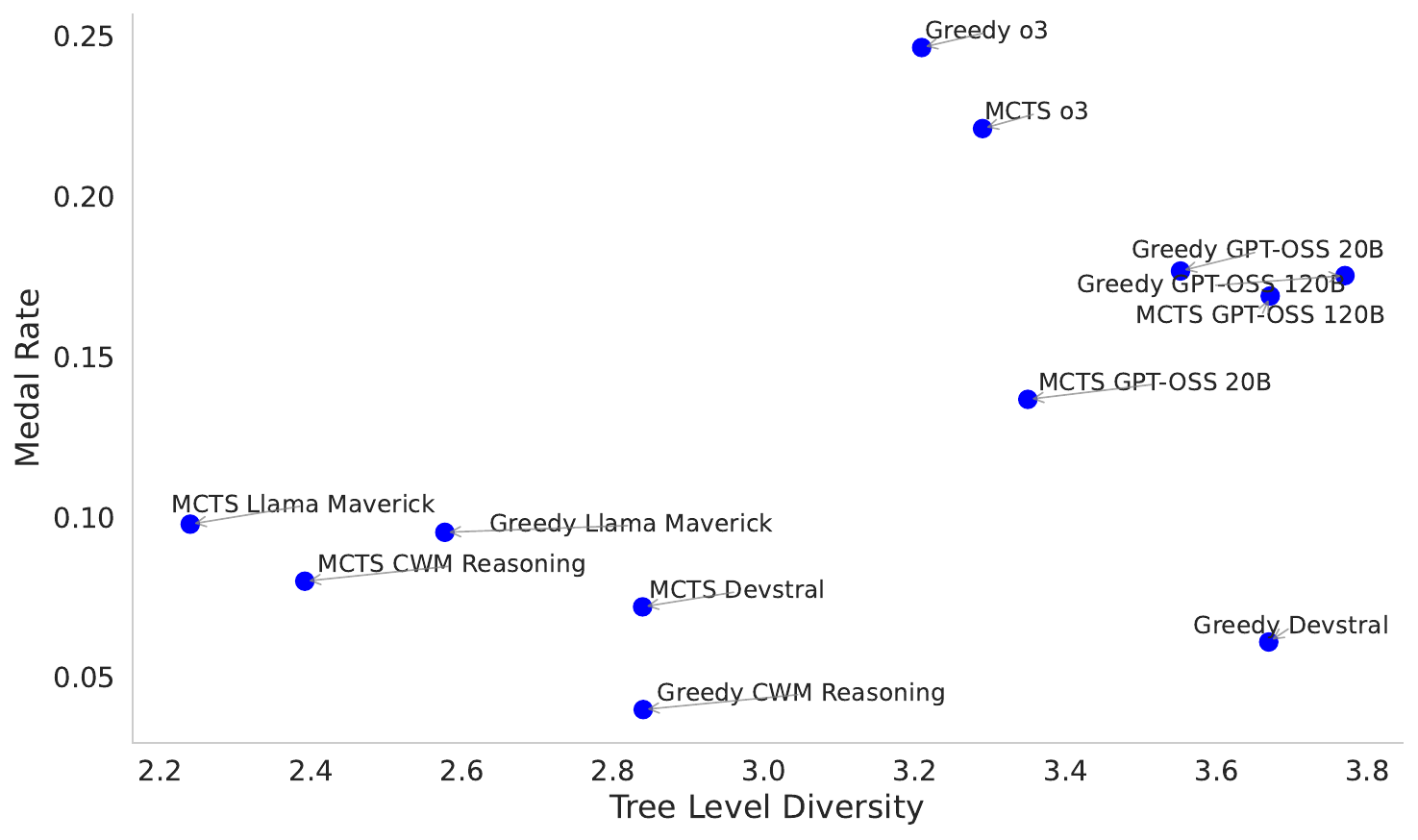}
    \caption{Correlation between tree-level diversity and performance on \mlebench}
    \label{fig:correlation_tree_level_diversity_performance}
\end{figure}

We observe a good correlation between diversity and performance, with two distinct clusters, one including high-performing  agents (using o3, gpt-oss 120b, and gpt-oss 20b as backbones) and the other using open-source LLMs in our study (Llama Maverick, Devstral, CWM).
Agents that utilize a wider range of techniques tend to achieve higher performance. Additionally, we observe in Figure~\ref{fig:correlation_tree_level_diversity_performance} how diversity changes for different agents, by measuring how many model architectures on average are used in the first 5 nodes of the agent's exploration, a metric we refer to as tree-level diversity.

Figure~\ref{fig:correlation_tree_level_diversity_performance} shows how high-performing models considered here (o3, gpt-oss 120b, gpt-oss 20b) use more diverse architectures in the 5 initial ideas (3.5 distinct architectures on average) compared to Llama Maverick, Devstral and CWM (2.8 distinct architectures on average).
Like diversity measured as entropy, tree-level diversity also correlates with performance. 

\subsection{Impact of Diversity: A Controlled Experiment}

\label{sec:results_controlled_exp}

We have seen that better agents are usually having more diversity of ideas in their trajectories. To understand whether diversity has a causal relationship with performance, we perform a controlled experiment, as described in Section~\ref{sect:control_div}, by removing the different mechanisms for diversity, and directly prompting the agents to generate similar ideas to solve a single task.

\begin{figure}[H]
    \centering
    \includegraphics[width=8cm]{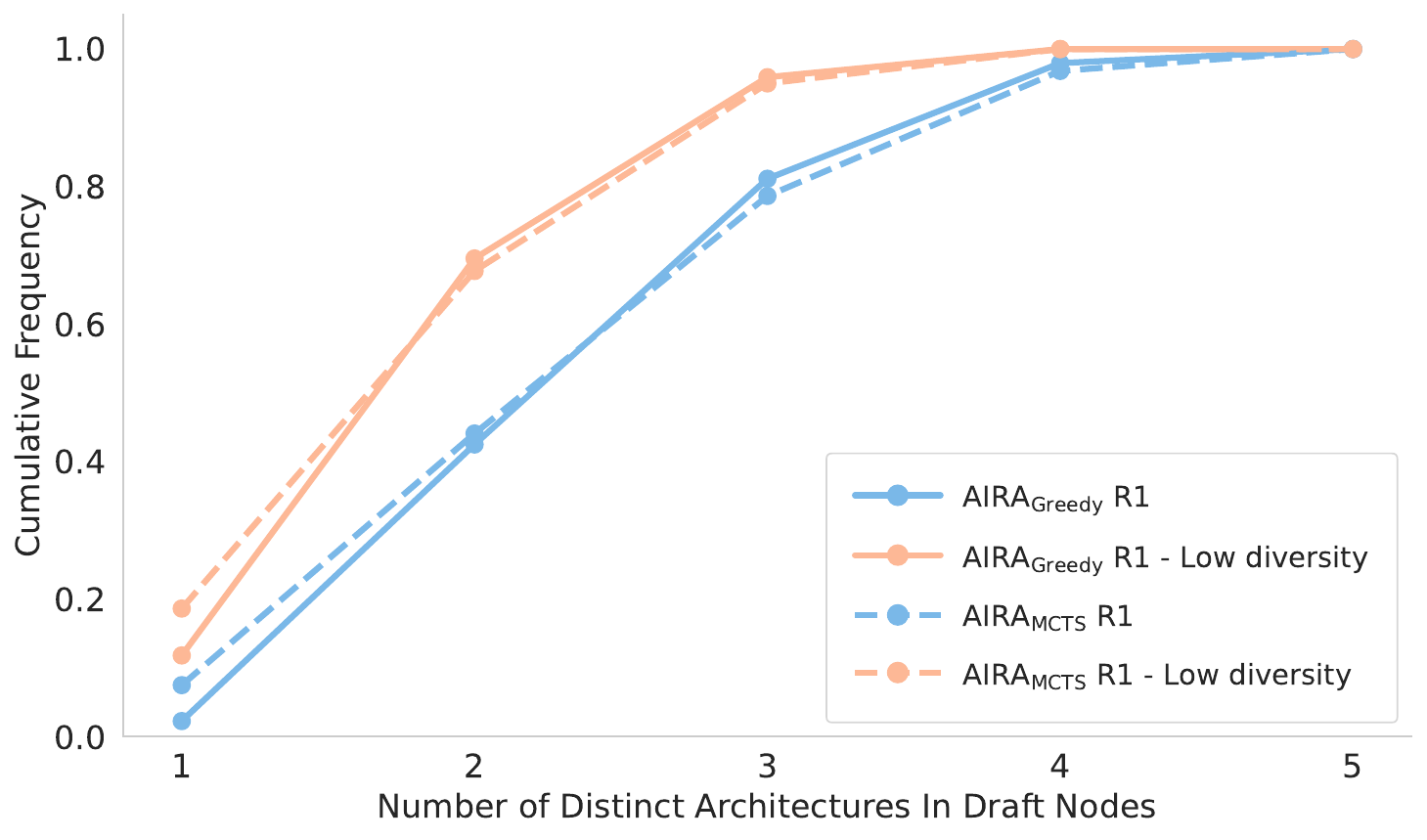}
    \caption{Number of distinct architectures per task - Cumulative Distribution}
    \label{fig:tree_level_diversity_experiment}
\end{figure}

\subsubsection{Are We Actually Influencing Diversity?}

Prompting the agent to come up with similar ideas negatively impacts the diversity of ideas generated by the agent. Figure  ~\ref{fig:tree_level_diversity_experiment} shows that agents with less diversity use a decreased number of unique architectures and general ML approaches.   
Baseline agents $\text{AIRA}_\text{Greedy}$ and $\text{AIRA}_\text{MCTS}$ use no more than 2 different architectures in their 5 initial drafts in only 40\% of tasks. However, the $\text{AIRA}_\text{Greedy} \text{- Low Diversity}$ and $ \text{AIRA}_\text{MCTS} \text{- Low Diversity}$ agents that are prompted to come up with similar ideas, use no more than 2 distinct architectures or approaches in 70\% of tasks. These different behaviors highlight the actual impact of diversity mechanisms on ideation.

\subsubsection{Results}

\begin{figure}[h!]
    \centering
    \includegraphics[width=8cm]{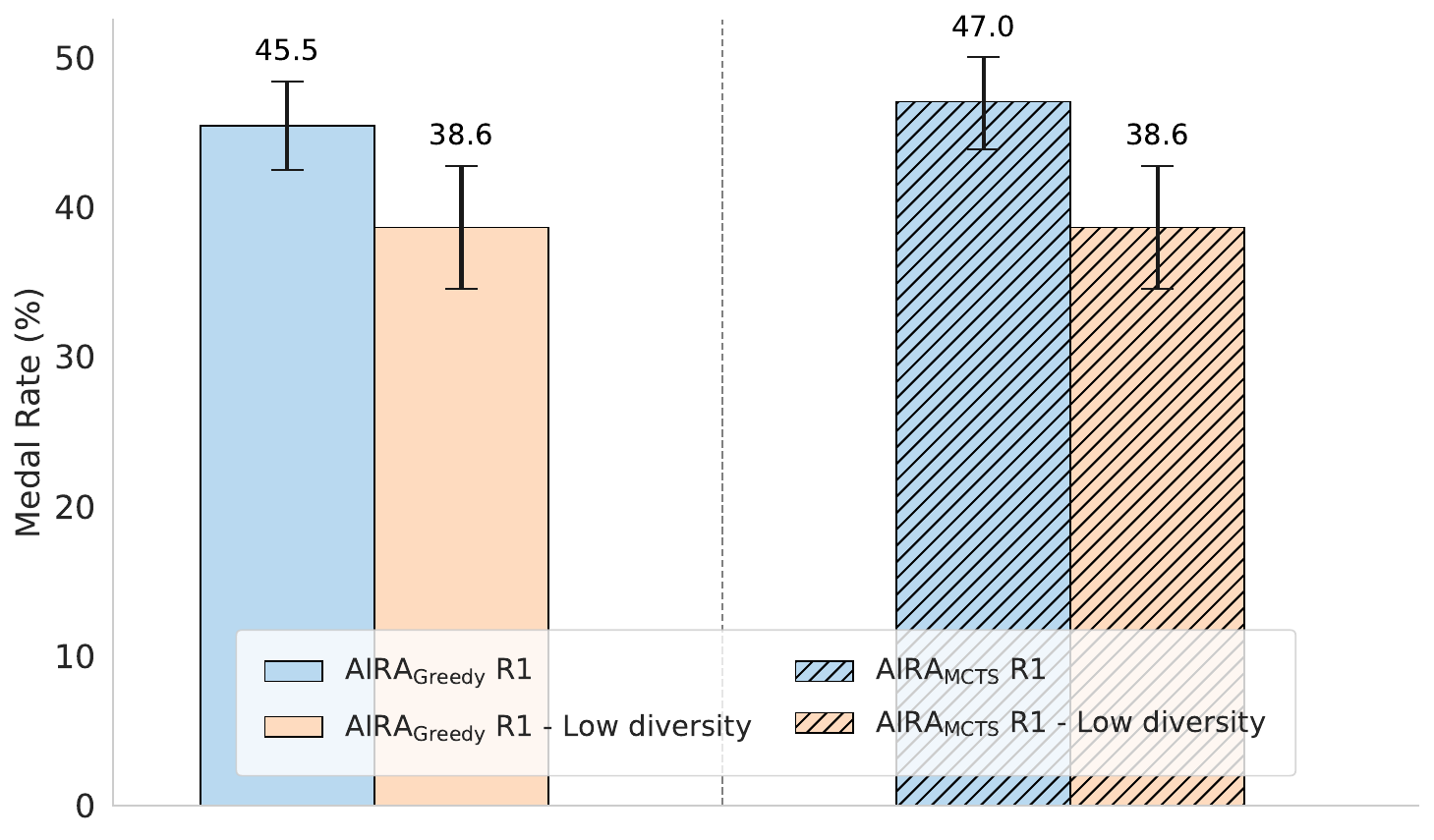}
    \caption{Comparison of \mlebench \space lite medal rate of $\text{AIRA}_\text{Greedy}$ and $\text{AIRA}_\text{MCTS}$ with and without interventions to reduce solution diversity (as indicated by '- Low diversity'). Error bars represent 95\% confidence intervals computed using stratified bootstrapping, using the rliable library  \citep{agarwal2021deep}.}
    \label{fig:controlledexperiment}
\end{figure}

When it comes to performance measured as medal rate, Figure~\ref{fig:controlledexperiment} demonstrates that reducing ideation diversity - by prompting the agent differently in order to generate similar ideas - leads to a decline of performance on \mlebench \space lite. This applies for both agentic scaffolds $\text{AIRA}_\text{Greedy} $ and $\text{AIRA}_\text{MCTS}$, with a $6.9$ and $8.4$ absolute points decrease, respectively. By modifying the system prompt to isolate the effect of ideation diversity, the results indicate that diversity is an important factor limiting performance.

\subsection{Evaluating with Alternative Metrics}
\label{sec:results_metrics}

While relevant, the differences in performance observed so far are constrained to the medal rate metric, used by default. The medal rate metric, by itself, may not offer a comprehensive picture of agent performance on \mlebench. In the current section, we introduce alternative metrics to provide a more complete assessment of performance on \mlebench.
\subsubsection{Alternative Metrics}

Each of the alternative metrics we consider offers a distinct perspective: while some, like medal rate, emphasize marginal improvements, others account for all performance gains. Additionally, certain metrics are based entirely on human score distributions, whereas others operate independently of them. We consider 4 additional metrics. 
\begin{enumerate*}
    \item \textbf{Valid Submission Rate}: The percentage of tasks in which the agent is able to make a valid submission. This metric captures the ability of the agent to ideate, implement, and debug until reaching at least one valid submission.
    \item \textbf{Average Normalized Score}: For each agent attempt at a task, we compute a normalized score: a score of 0 represents the lowest human score achieved on the task, and 1 the highest. 
    The metric captures how good agents submissions are, independently from human score distributions.

    \item \textbf{Percentile}: The metric captures the ability of the agent to outperform humans at machine learning engineering. Compared to medal rate, this metric still relies on human score distributions, and offer a less discrete assessment of the performance. Like average normalized scores, and unlike medal rates, improvements of percentile in poor and strong scores are equally valued.

    \item \textbf{ELO-Based Agent Ranking}: We create an ELO system~\citep{elo} using all possible heads-to-heads between agents' scores. ELO rankings are agnostic of the human score distribution on \mlebench \space tasks. ELO difference of 100 points corresponds to about a 64\% expected win probability for the higher-rated agent.
\end{enumerate*}

\subsubsection{Data Analysis Results with Additional Metrics}
\label{newmetrics}

\begin{figure}[h]
  \centering
  \begin{minipage}{0.45\textwidth}
    \centering
    \includegraphics[width=\linewidth]{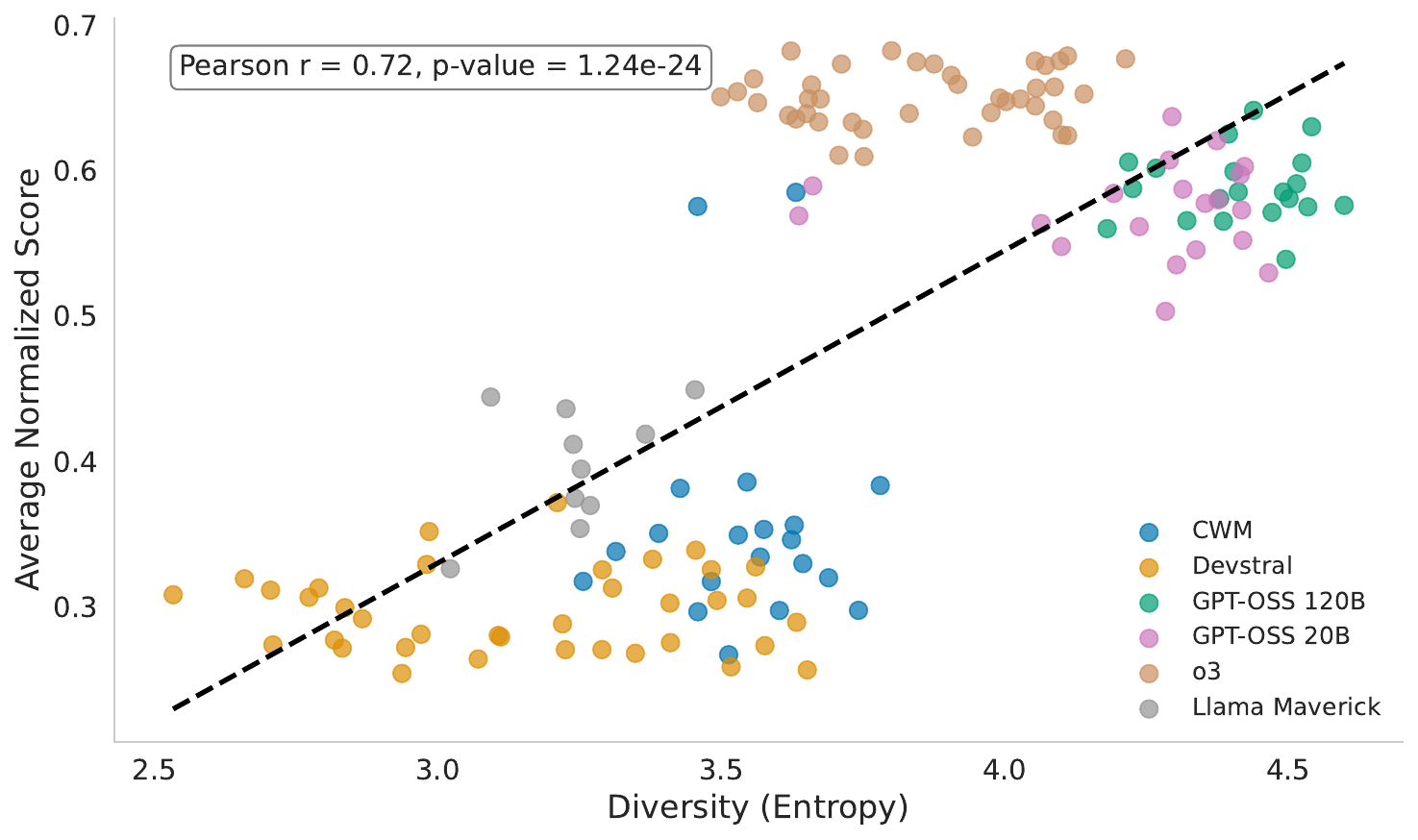}
    \caption{Correlation between diversity and performance measured as average normalized score}
    \label{fig:correlationdiversityaveragescore_perseed}
  \end{minipage}
  \hfill
  \begin{minipage}{0.45\textwidth}
    \centering
    \includegraphics[width=\linewidth]{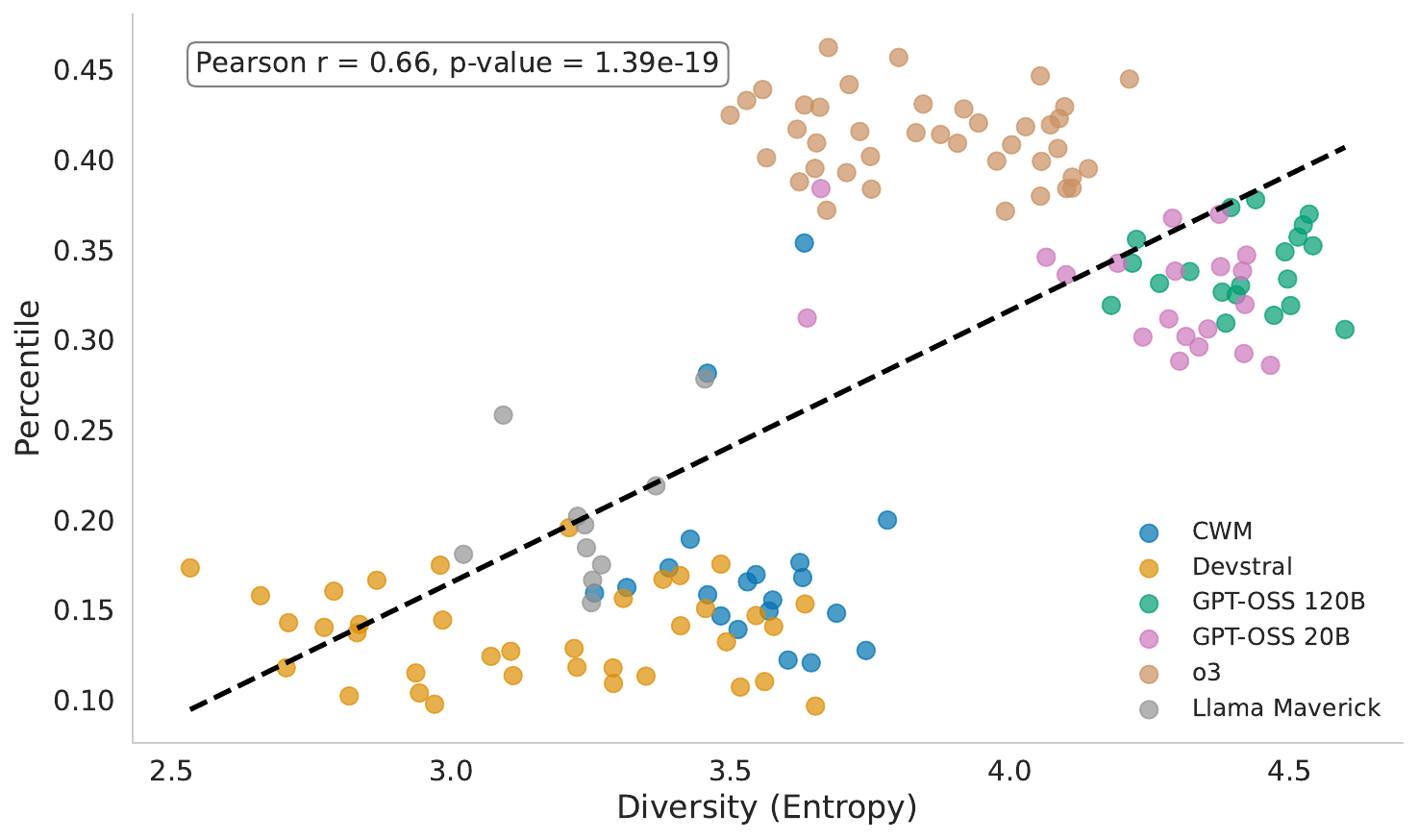}
    \caption{Correlation between diversity and performance measured as percentile}
    \label{fig:correlationdiversitypercentile_perseed}
  \end{minipage}
\end{figure}

We use the new set of metrics to gain a deeper understanding of the correlation between diversity and performance. When measuring performance using either the percentile or the average normalized score, instead of the medal rate, our correlation results remain consistent and, in fact, show even higher correlations (Figures~\ref{fig:correlationdiversityaveragescore_perseed} and~\ref{fig:correlationdiversitypercentile_perseed}).

\begin{figure*}[htbp]
    \centering
    \includegraphics[width=0.75\textwidth]{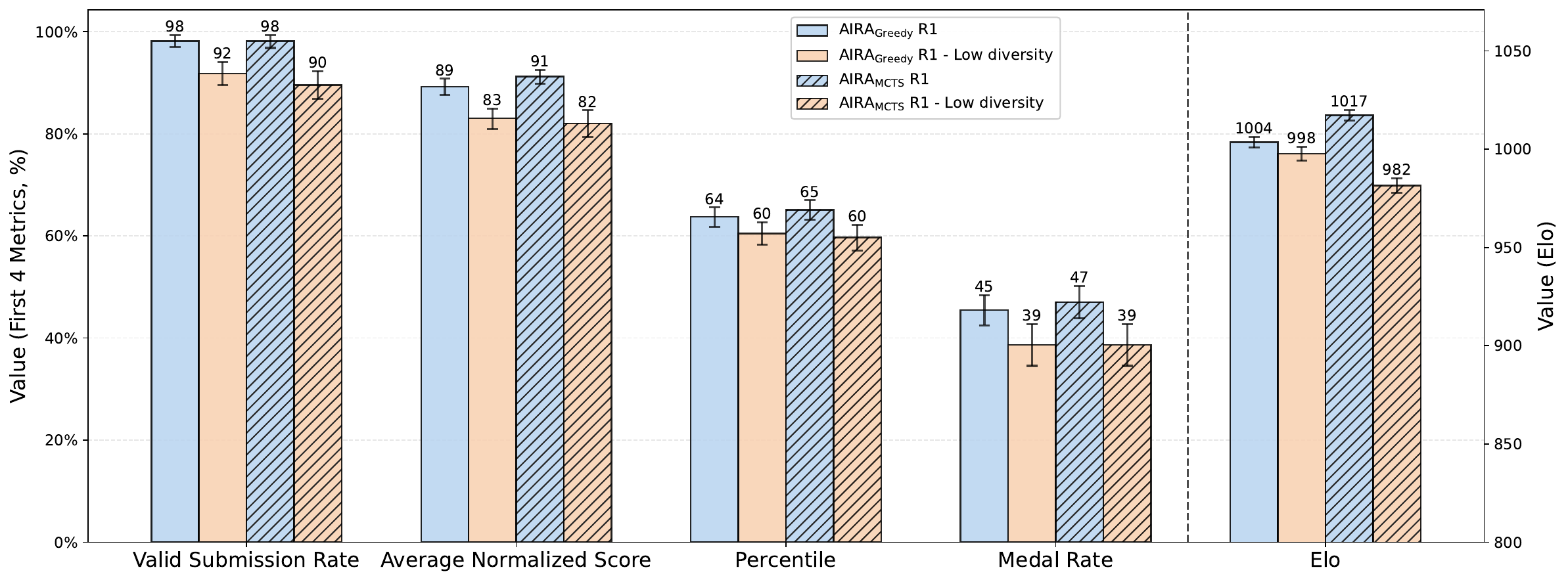}
    \caption{Results of the Controlled Experiments with Additional Metrics}
    \label{fig:Experiment Results_Additional Metrics}
\end{figure*}

\subsubsection{Controlled Experiment Results with Additional Metrics}

    Using this additional set of metrics, we can perform a more comprehensive analysis of the controlled experiment. Figure~\ref{fig:Experiment Results_Additional Metrics} shows the performance gap between baseline agents ($\text{AIRA}_\text{Greedy}$, and $\text{AIRA}_\text{MCTS}$) and agents with ablated diversity. The performance loss is observed across all different metrics, validating our hypothesis that agents perform worse when ideas are less diverse.

A notable observation is the drop in valid submission rates, which fall from 98\% to 92\% for $\text{AIRA}_\text{Greedy}$ and to 90\% for $\text{AIRA}_\text{MCTS}$. This indicates that, for certain tasks, Low Diversity agents are sometimes unable to produce even a single valid submission during their search. Our analysis reveals that this decline is primarily driven by two competitions: 'text-normalization-challenge-english-language' and 'text-normalization-challenge-russian-language'. Upon examining agent trajectories, we find that Low Diversity agents repeatedly attempt to implement the same model, T5 \citep{t5}, but consistently fail, resulting in timeouts. In contrast, baseline agents implement a wider range of solutions and are more often able to make correct submissions. Notably, baseline agents also occasionally attempt to implement T5 and encounter similar failures, but their greater ideation diversity allows them to succeed elsewhere.
These two competitions also account for a significant portion (estimated at around half) of the observed decline in medal rates.

Our findings suggest that one reason for the drop in performance is that Low Diversity agents tend to focus on similar ideas that they sometimes cannot successfully implement. In other words, ideation diversity is crucial for performance, as it increases the likelihood that an AI research agent will attempt solutions it is actually capable of executing.

\section{Discussion}
\label{sec:discussion}

\paragraph{What does it take to be a good AI research agent?} We can imagine a hypothetical future scenario where excellent AI research agents  ideate brilliant experiments and have outstanding coding skills to implement them. Until we get to this ideal situation, in practice, even state-of-the-art AI research agents will exhibit limited ideation and implementation capabilities, particularly when  evaluated in challenging, real-world settings. In this imperfect, yet realistic, scenario, given the same level of capabilities, we prefer agents with greater ideation diversity. First, because it de-risks implementation pitfalls. Our analysis of the controlled experiment shows that one reason why diversity is important is that it helps agents design solutions they are actually able to execute, highlighting the interplay between ideation and implementation. If the different proposed plans by the agent rely on similar approaches, and those happen to be hard to implement by the agent (in the context of the particular task), then we risk low implementation accuracies. Intuitively, a potential second argument for ideation diversity is that given the difficulty of coming up with creative, yet feasible research ideas, exploring significantly different paths hedges against pursuing a single unproductive direction (even if the agent knows how to implement it), and enables agents to more
effectively explore the solution space of machine learning problems. We want to invest the allocated compute in a diversified, yet plausible, set of ideas. However, this second reason is hard to evaluate given the implementation bottleneck. Ultimately, a good experimentation plan could fail due to the agent being unable to implement it. Repeating these controlled experiments as LLMs' coding capabilities get increasingly more powerful may yield valuable insights.

\begin{figure}[h]
  \centering
  \begin{minipage}{0.45\textwidth}
    \centering
    \includegraphics[width=\linewidth]{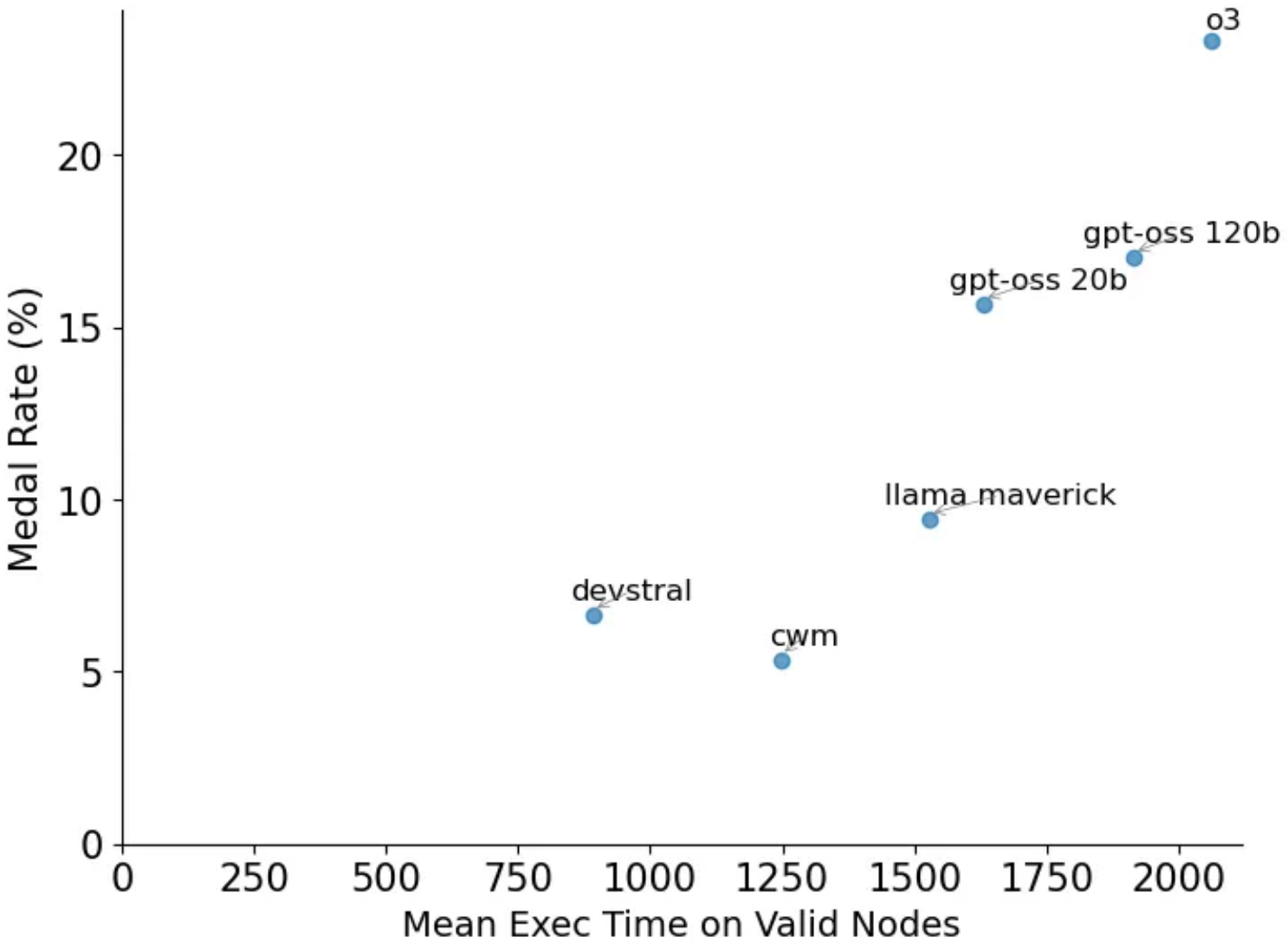}
    \caption{Correlation between the average execution time on valid nodes and performance (\mlebench, 75 tasks)}
    \label{fig:correlation_meanexectime_performance}
  \end{minipage}
  \hfill
  \begin{minipage}{0.45\textwidth}
    \centering
    \includegraphics[width=\linewidth]{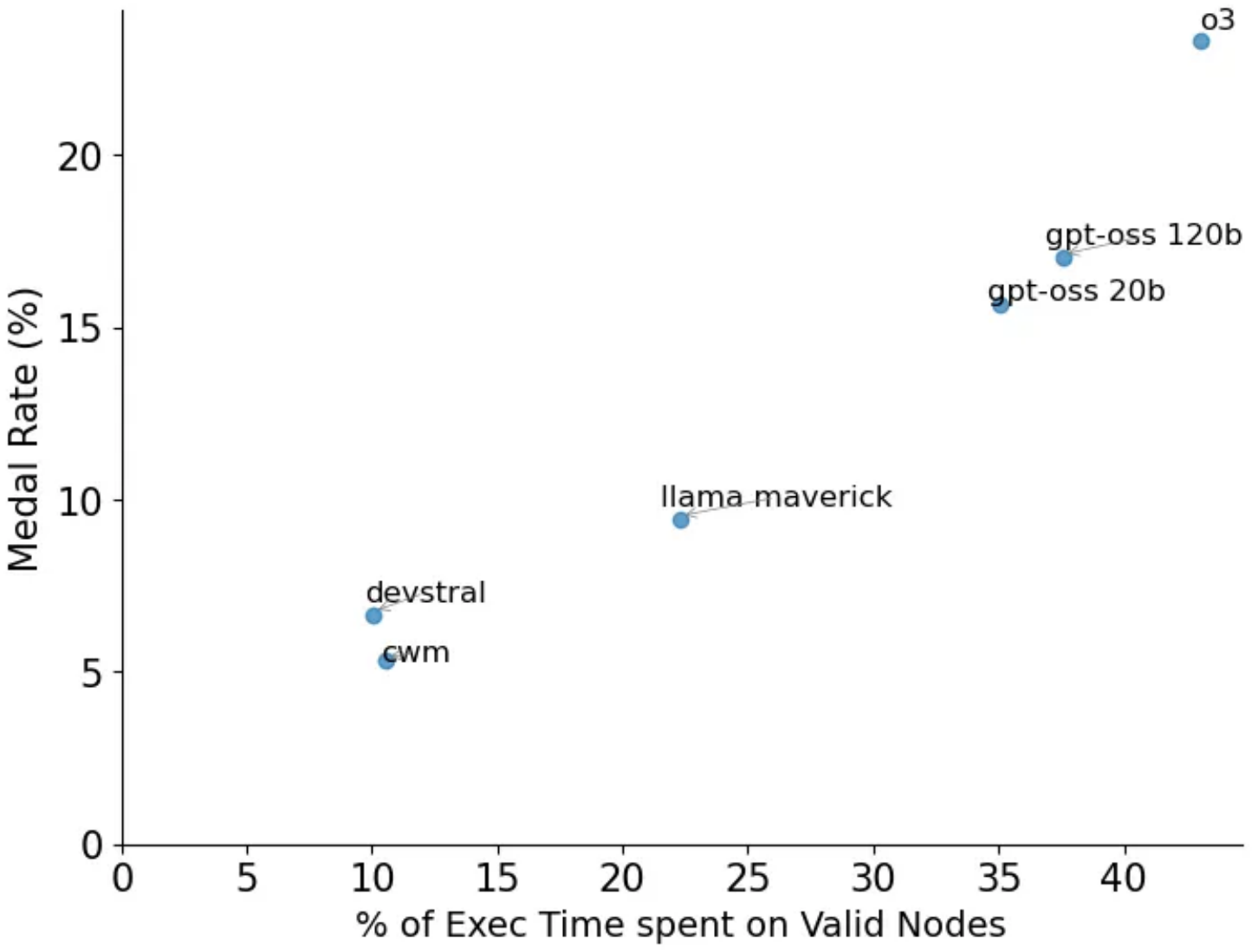}
    \caption{Correlation between the share of execution time spent on valid nodes and performance (\mlebench, 75 tasks)}
    \label{fig:correlation_shareexectime_performance}
  \end{minipage}
  
\end{figure}

\paragraph{Importance of the implementation bottleneck.} 
Unsurprisingly, implementation quality is an important bottleneck of AI research agents.
We observe a strong correlation between AI research agents' performance and the ability to implement sufficiently complex solutions. By aggregating performances of AIRA (Greedy and MCTS) 
for each LLM, Figure~\ref{fig:correlation_meanexectime_performance} shows that, on average, the more time an agent spends on each successfully implemented solution (including ideation, implementation, and model training), the more medals it earns. This suggests that performance increases with the agents' ability to implement more complex solutions.
Furthermore, Figure~\ref{fig:correlation_shareexectime_performance} shows that agents perform better when, out of the 24 hours allotted to complete a task, they spend a higher proportion of time on successfully implemented solutions.
However, since LLMs and coding agents are improving rapidly \citep{kwa2025measuringaiabilitycomplete}, particularly in verifiable tasks \citep{deepseekai2025deepseekr1incentivizingreasoningcapability}, we hypothesize  that the relative importance of the ideation and planning phase might increase over time, not to de-risk implementation pitfalls, but to efficiently explore the solution space.

\paragraph{Generalization to other benchmarks.}
The findings presented in this study are based on experiments conducted using \mlebench \space only. Given the range of machine learning tasks included in this benchmark, we hypothesize that our results are likely to generalize to other machine learning tasks. Additional benchmarks could be examined in future research.

\paragraph{Limitations of \mlebench \space evaluations.} Performance on \mlebench \space has traditionally been evaluated using Kaggle’s medal system, where medals are awarded based on score percentiles. For example, in competitions with fewer than 99 teams, gold medals are given to the top 10\% of submissions. However, this medal-based evaluation framework has several important limitations.

  First, medal criteria change with the number of submissions (e.g., bronze goes to the top 10\% for 1000+ teams, but top 40\% for 1–249 teams), so earning a medal does not indicate consistent performance across competitions. Second, as shown in the Appendix, the gap between the bronze threshold and the best score is often extremely small—frequently below 3\%. 
  
  Since agents are evaluated on custom test sets, while medal thresholds are computed using Kaggle private test sets, the data split variance introduces some score variability, and can affect medal outcomes.
Third, some competitions are over a decade old, and human score distributions from these may not represent current machine learning standards. In some older competitions like 'detecting-insults-in-social-commentary', AIRA agents are able to outperform best human submissions. In recent competitions (post-2022), agent performance drops sharply, with most agents unable to earn medals. To address these different issues, we also used additional metrics, in Section \ref{sec:results_metrics}. In the Appendix, we provide additional information on these alternative metrics.

\paragraph{Limitations of this study.} Despite the efforts to isolate ideation diversity, it is difficult to track the potential second-order effects of modifying the system prompt. To better isolate the effect of ideation diversity, future work could focus on disentangling the LLM responsible for ideating, and the one responsible for implementing. In this work, we also experimented with controlling diversity through temperature, as detailed in the Appendix.

\section{Related Work}
\label{sec:related}
\paragraph{Generation diversity in language models.} Generation diversity in language models has been studied and even explicitly promoted since the statistical machine translation era~\citep{macherey-och-2007-empirical,gimpel-etal-2013-systematic, xiao2013bagging}, where selecting~\citep{devlin-matsoukas-2012-trait} or combining~\citep{macherey-och-2007-empirical} a set of diverse yet plausible generations lead to improved translation quality.  
 Similar observations were later identified in neural machine translation and other sequence-to-sequence settings~\citep{li2015diversity,vijayakumar2016diverse,ippolito2019comparison}. \citet{holtzman2020curiouscaseneuraltext} developed nucleus sampling with the goal of generating both coherent and diverse text. More recently, \citet{Murthy_2025} and \citet{DBLP:conf/iclr/KirkMNLHGR24} study the effect of RLHF~\citep{ouyang2022traininglanguagemodelsfollow} on LLMs with a focus on the (decreased) generation diversity. \citet{chen2024diversitysyntheticdataimpact} study effect of diversity of synthetic data in training LLMs. \citet{li2025preserving} propose a diversity-preserving algorithm for supervised fine-tuning of LLMs. 

\paragraph{Diversity in reinforcement learning and population-based reinforcement learning.} Trajectory diversity is synonym of increased exploration in reinforcement learning models. \citet{hong2018div-driven} investigate a diversity-driven exploration strategy for training reinforcement learning models. \citet{eysenbach2018diversity} propose a diversity objective to learn skills without reward functions. 
\citet{holder2020} improve diversity in population-based reinforcement learning, while \citet{conti2018}  propose a novelty objective in a population of agents to improve exploration in reinforcement learning. 
More recently, \citet{yao2025diversityawarepolicyoptimizationlarge} propose a diversity-aware policy optimization algorithm; unlike the works cited above, it does so in the context of LLMs. \citet{zeng2025bstar} propose B-star, a reinforcement learning approach for reasoning LLMs  that balances exploration and exploitation.

\paragraph{Diversity in multi-agent systems.} Diversity has also been studied in the context of multi-agent foundation models~\citep{tuyls2023}. \citet{bettini2025impactbehavioraldiversitymultiagent} study the impact of behavioral diversity in multi-agent reinforcement learning. \citet{li25aamasdiversity}
 address how agents in multi-agent RL often end up learning nearly identical behaviors when they share network parameters, which hurts exploration. The authors propose CTEM, a method that uses trajectory entropy maximization to push agents toward more diverse behaviors without needing complex density models.
 \citet{li2025llmassisted} find that generating diverse teammates in multi-agent training can lead to random, semantically meaningless behaviors, reducing training efficiency. Their SemDiv approach leverages large language models to describe coordination strategies in natural language, then converts these into reward functions for training meaningful teammate policies.
In LLM-agent world simulations, \citet{chu2025exploringcontrollingdiversityllmagent} investigate prompt design’s impact on conversational diversity and introduce a prompt-tuning mechanism that controls diversity via a single parameter.

\paragraph{Automated machine learning and AI research agents.} In the era of tool-use LLM-based agents~\citep{schick2023toolformer, kaddour2023challengesapplicationslargelanguage}, \citet{shen2023hugginggptsolvingaitasks, nathani2025mlgymnewframeworkbenchmark} evaluate agents at implementing machine learning tasks. \mlebench~\citep{chan2025mlebench}, the benchmark used in this work, is a machine learning benchmark consisting of 75 Kaggle tasks. \citet{zhao2025automatedllmspeedrunningbenchmark} propose a self-contained machine learning task with a focus on language models. AIRA \citep{toledo2025airesearchagentsmachine}, studies AI research agents by formalizing AI research as search policies over a space of candidate solutions.  This development is parallel to agentic benchmarks and scaffolds for other fields, such as software engineering~\citep{jimenez2024swebench, jiang2025aideaidrivenexplorationspace}.

\section{Conclusions}
\label{sec:conclusions}

This work started from the hypothesis that ideation diversity is a key bottleneck in AI research agents' performance. We have confirmed this hypothesis by conducting a large-scale analysis on AI research agents' trajectories and performing a controlled experiment. Our findings hold across several evaluation metrics. 

In future work, we suggest focusing on diversity-aware methods, as other bottlenecks such as implementation quality will decrease in importance when AI systems keep improving. We also recommend considering multiple evaluation metrics and extending the existing benchmarks to more recent machine learning tasks.

\bibliography{main}
\bibliographystyle{assets/plainnat} 

\clearpage
\appendix
\onecolumn
\section{Appendix}
\label{sec:appendix}

\subsection{Controlling Ideation Diversity Using Temperature}

With the intuition that the temperature sampling parameter has an impact on ideation diversity and performance, we run several experiments intervening on the sampling temperature. To isolate the impact of temperature, we use $\text{AIRA}_\text{Greedy}$ but remove all mechanisms that enhance ideation diversity (sibling memory, prompt-adaptive complexity, mention of diversity in the system prompt), as described in section~\ref{sect:control_div}, and experiment with different temperatures, above and below the recommended temperature of 0.6 for DeepSeek-R1~\citep{deepseekai2025deepseekr1incentivizingreasoningcapability}.

\begin{figure}[H]
    \centering
    \includegraphics[width=14cm]{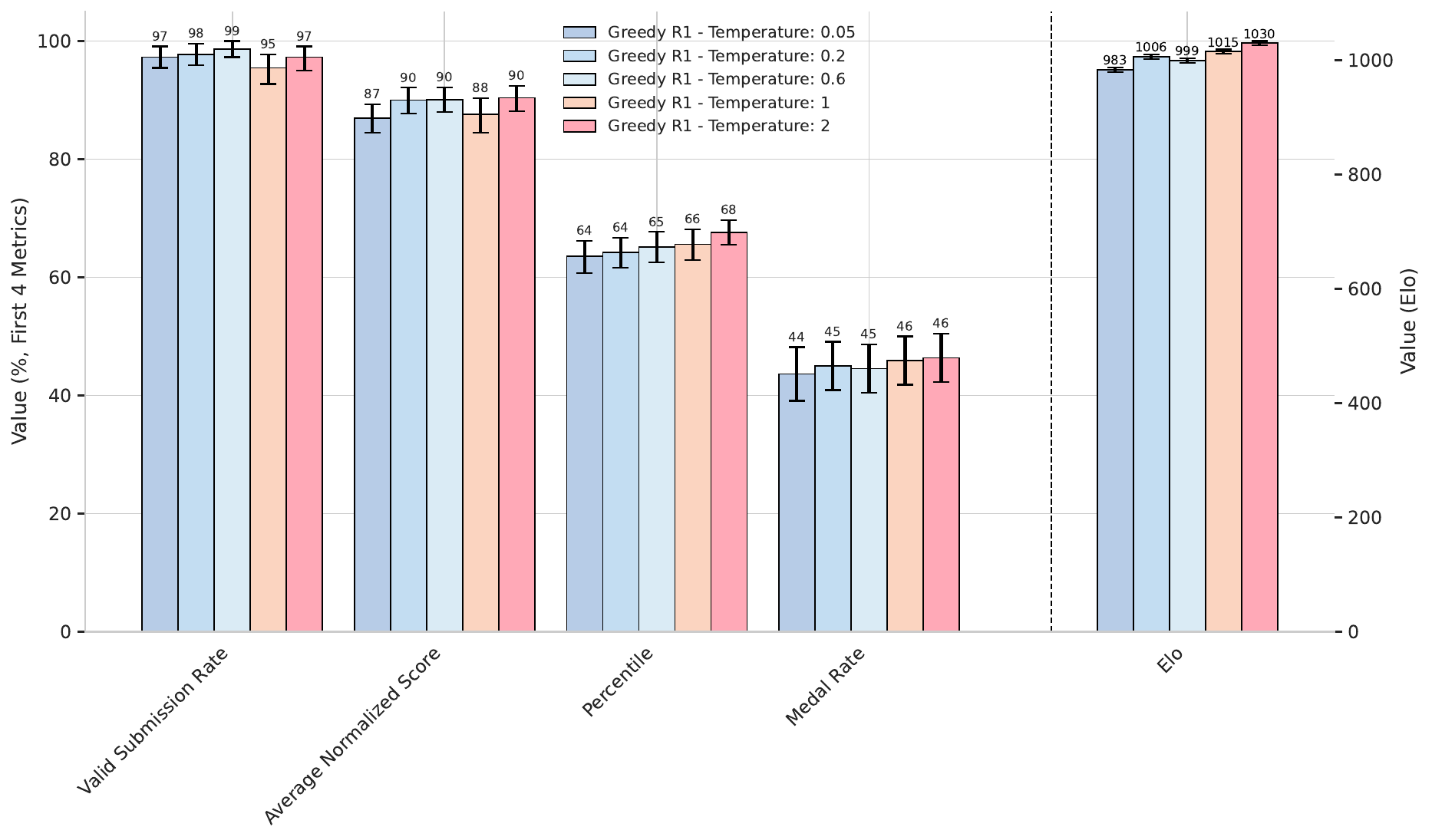}
    \caption{Agent Performance With Different Temperature Settings}
    \label{fig:temperatureexps}
\end{figure}

Figure~\ref{fig:temperatureexps} shows that changing temperature does not have an impact on performance (neither beneficial nor detrimental), assessed as medal rate. Results hold across alternative metrics like valid submission rate, average normalized score, and percentile, except for Elo, the only metric where increased temperature significantly leads to improved performance.  Despite our effort in isolating ideation diversity, we hypothesize that the reason for these negative results is that modifying the temperature parameter (instead of using the recommended one) also affects the agent in additional manners \citep{Renze_2024,wu2025roletemperaturesamplingtesttime,li2025exploringimpacttemperaturelarge}. For example, we would expect the implementation capabilities to also be affected,\footnote{Surprisingly, \citet{Renze_2024} found no statistically significant effect in problem solving skills of LLMs for certain temperature ranges.} and there could be second-order effects that are hard to reason about. We leave further investigating the temperature-based results as future work. 

\label{other_metrics_appendix}

\subsection{Alternative Metrics}

\subsubsection{Limitations of MLE-Bench Medal System}

Performance on \mlebench \space has traditionally been evaluated using Kaggle’s medal system, where medals are awarded based on score percentiles (see details here). For instance, in competitions with fewer than 99 teams, gold medals are given to the top 10\% of submissions. However, several limitations are inherent to this medal-based evaluation framework.

\begin{figure}[h]
  \centering
  \includegraphics[width=0.45\linewidth]{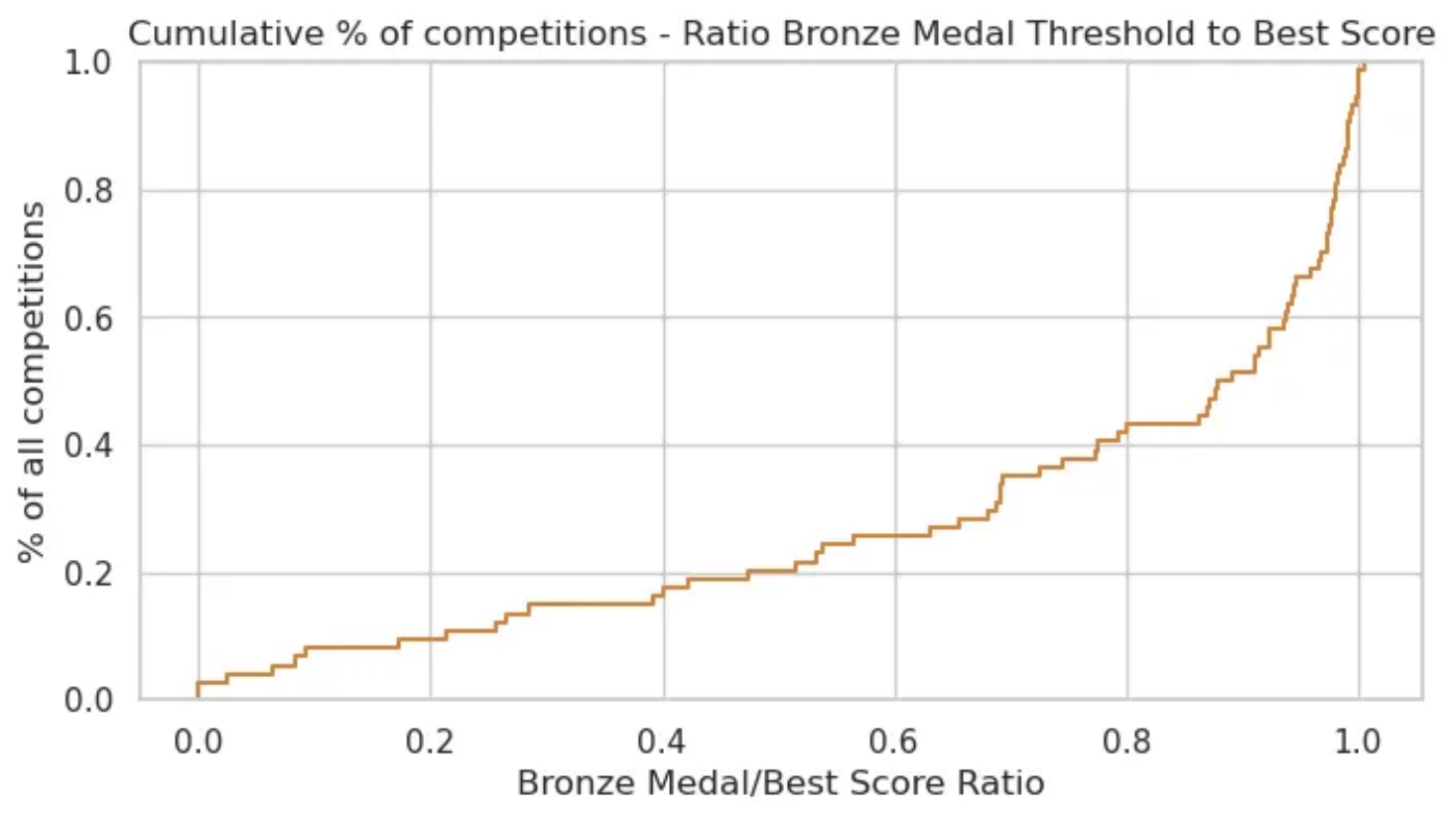}
  \caption{Ratio between the bronze medal threshold and the best score for each competition on \mlebench. In many cases, the ratio is close to 1, indicating a very narrow margin between the bronze threshold and the top score.}
  \label{fig:bronze_to_best_ratio}
\end{figure}

\paragraph{Variable medal thresholds} Kaggle medal criteria vary with the number of submissions (e.g., bronze is the top 10\% for competitions with 1000+ teams vs. top 40\% for those with 1-249 teams; see Figure \ref{fig:bronze_to_best_ratio}). This inconsistency means that medals do not equate to the same performance level across different competitions.

\begin{table}[ht]
\centering
\begin{tabular}{l|c|c|c|c}
& \textbf{0 - 99 teams} & \textbf{100 - 249 teams} & \textbf{250 - 999 teams} & \textbf{1000+ teams} \\
Bronze & Top 40\% & Top 40\% & Top 100 & Top 10\% \\
Silver & Top 20\% & Top 20\% & Top 50 & Top 5\% \\
Gold & Top 10\% & Top 10 & Top 10 + 0.2\% & Top 10 + 0.2\% \\

\end{tabular}
\caption{Kaggle Competition Medals}
\label{tab:kaggle_medals}
\end{table}

    \paragraph{Narrow thresholds between medals and top scores} In some competitions, there is only a minimal difference between the threshold for earning a medal and the highest score achieved. Specifically, as shown in Figure~\ref{fig:bronze_to_best_ratio}, in about 30\% of \mlebench \space competitions, the score (relative) difference between the best score and the bronze medal threshold is less than 3\%. In around 50\% of all \mlebench \space lite competitions, this gap is similarly small.

    \paragraph{Different test sets for evaluation} AI research agents are tested on custom sets designed by MLE-bench, not the private test sets used for human submissions on Kaggle. Human submissions leading to medals or best scores would perform differently on custom test sets for AI research agents, while the medal thresholds are still computed based on submissions on Kaggle private test sets. This variability is an issue, especially when medal thresholds are set close to each other, and could make medal thresholds unreachable by AI research agents.
    \paragraph{Competition age} Some competitions are more than 10 years old. In this rapidly evolving space, human score distributions don't reflect what they would be now in 2025 (for example agents perform better than top human scores on 'detecting-insults' competition). For recent competitions, after 2022, we see a substantial decrease in performance, and all agents tested are not able to get any medal for most competitions after 2022.

These limitations suggest that the medal rate metric, by itself, may not offer a comprehensive evaluation of agent performance on \mlebench. Therefore, in the following section, we describe alternative metrics to provide a more complete assessment of performance on \mlebench. We used these metrics for the results in Section \ref{sec:results_metrics}.

\subsubsection{Evaluation Principles}
Evaluations serve as a measure of a model's capabilities. For \mlebench, we aim to evaluate our AI research agent’s ability to address a wide range of machine learning tasks, and manage key stages of the machine learning lifecycle in order to get the best performance, including data cleaning \& preprocessing, model development \& tuning, validation. We describe here a list of principles to consider for additional metrics.


\paragraph{Limited Set of Metrics} 
We have shown that relying solely on medal rates was hindering our assessment of performance on \mlebench. We need more metrics to get a comprehensive view of the performance of AI research agents. However, relying on a large set of metrics can make the assessment of an agent difficult, and hard to compare with others.
\paragraph{Inclusion of all attempts}
Failed submissions need to be taken into account in the evaluation. If an agent gets a perfect score on 50\% of tasks and fails to get a valid submission on the other 50\% of tasks, the agent should theoretically not get a perfect score.
\paragraph{Independence from Human Score Distributions} 
Agents and humans are not evaluated on the same test sets. 

Medal rates only value marginal improvements on good scores.
\paragraph{Capturing the complexity of hill-climbing}
In some competitions, the complexity resides in doing the last mile of optimization.

\begin{table}[H]
\resizebox{\textwidth}{!}{
\begin{tabular}{c|c|c|c|c}
\textsc{Metric} &\textsc{ Independent from human scores} & \textsc{Values all improvements} & \textsc{Inclusion of all attempts} & \textsc{Captures hill-climbing complexity}  \\
\hline
Valid Submission Rate & \xmark & \xmark & $\checkmark$ & \xmark \\
Medal Rate & \xmark & \xmark & $\checkmark$ & $\checkmark$ \\
Human Score Percentile & \xmark & $\checkmark$ & $\checkmark$ & $\checkmark$ \\
Average Normalized Score & $\sim$ (human score bounds) & \xmark& $\checkmark$ & \xmark \space (all improvements valued equally) \\
ELO-based ranking & $\checkmark$ & $\sim$ (values beating another agent) & $\checkmark$ & $\sim$ (depends on agents' score distribution) \\

\end{tabular}

}
\caption{Additional Metrics}\label{tab:add_metrics}
\end{table}

Table \ref{tab:add_metrics} shows the additional metrics used in this work and how they relate to these principles. None can actually follow all these principles at the same time, demonstrating the importance of having multiple metrics to describe the performance of AI research agents.

\subsection{Models Used By Agents For Image Classification Tasks}

\begin{figure*}[htbp]
    \centering
    \begin{minipage}[t]{0.48\textwidth}
        \centering
        \includegraphics[width=\linewidth]{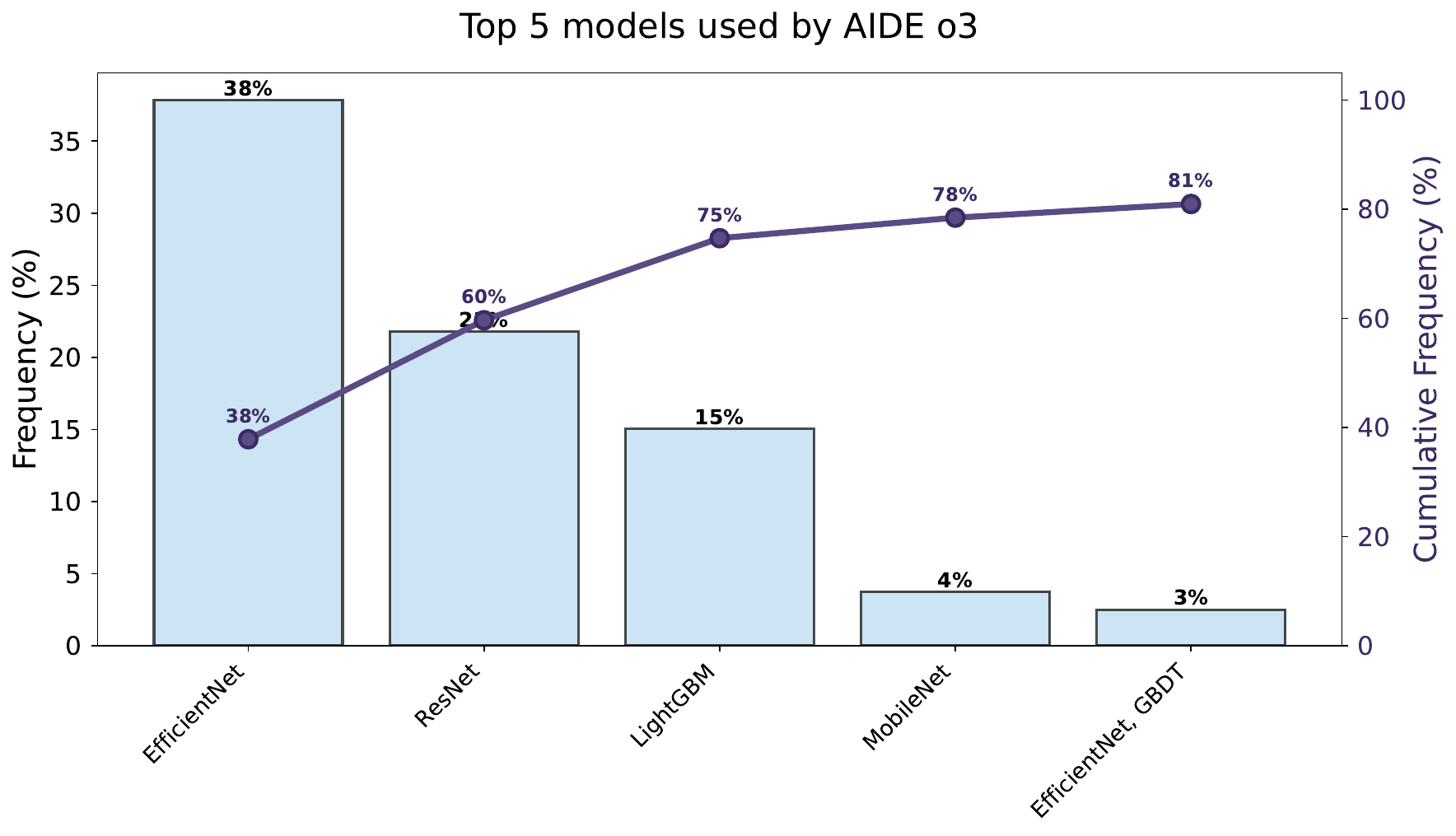}
        \caption*{(e) Diversity of models for Image Classification Tasks - AIDE}
    \end{minipage}%
    \hfill
    \begin{minipage}[t]{0.48\textwidth}
        \centering
        \includegraphics[width=\linewidth]{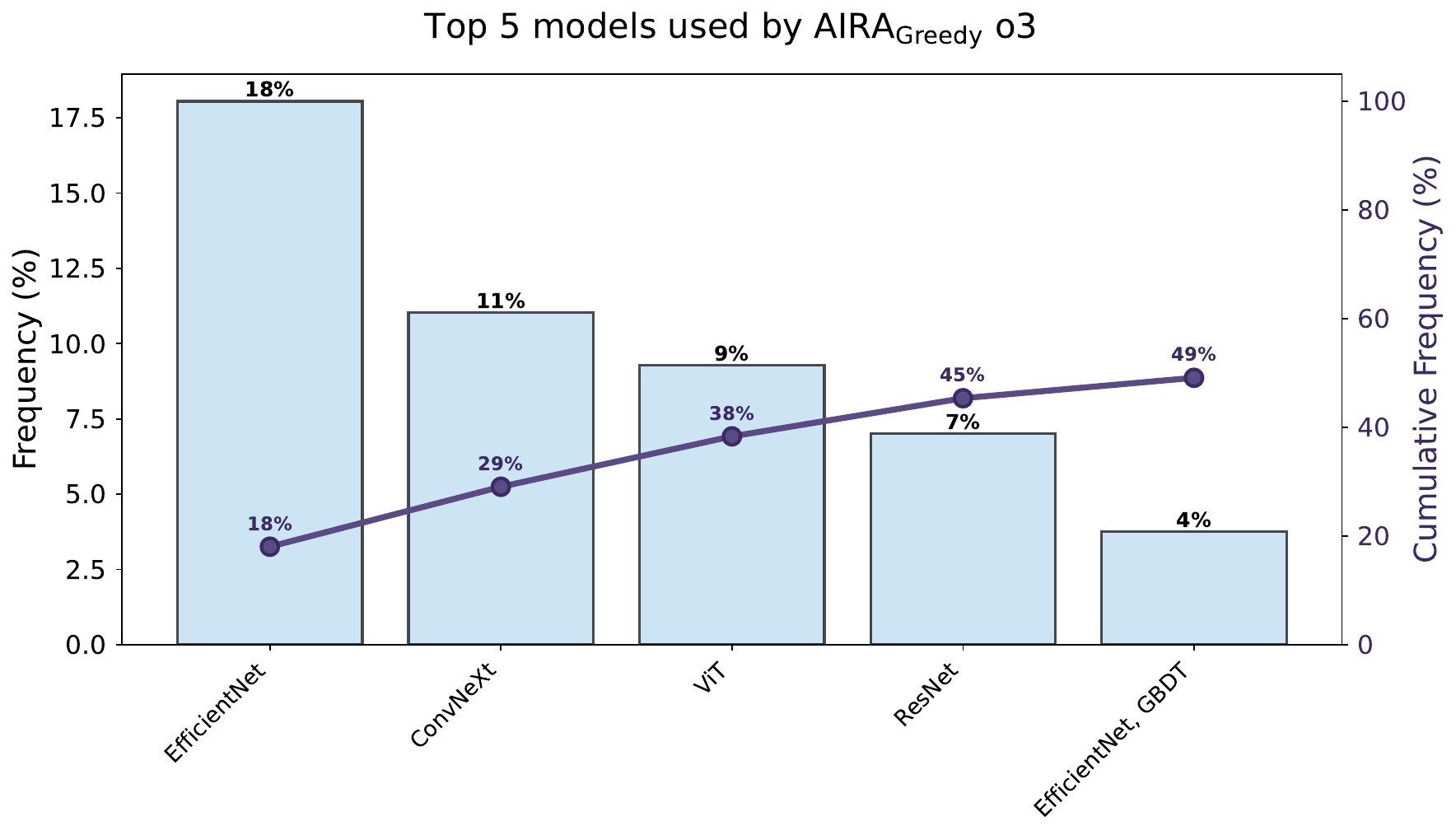}
        \caption*{(f) Diversity of models for Image Classification Tasks - $ \text{AIRA}_{\text{Greedy}} $}
    \end{minipage}
    \caption{Models used by AIDE and $ \text{AIRA}_{\text{Greedy}} $ for image classification tasks.}
    \label{fig:image_classification_models}
\end{figure*}

Focusing on the neural architectures used by the agents in image classification tasks in Figure~\ref{fig:image_classification_models},\footnote{Image classification is the largest category in \mlebench \space (8 tasks out of 22 in \mlebench \space lite)} we observe that AIDE relies on the EfficientNet architecture (and its variants) for almost 40\% of the tasks. For 75\% of its attempts, AIDE uses only 3 different architectures: EfficientNet, ResNet \citep{he2016deep} and LightGBM.
$\text{AIRA}_\text{Greedy}$ uses a wider range of architectures, with EfficientNet, ConvNeXt \citep{liu2022convnet}, and ViT \citep{dosovitskiy2020image} making only 38\% of the agents' initial ideas.

\end{document}